\documentclass{WileyMSP-template}

\usepackage{amsfonts}
\usepackage{dsfont}
\usepackage{stfloats}
\usepackage{booktabs}
\usepackage{pifont}
\usepackage{geometry}
\usepackage[inkscapelatex=false]{svg}
\usepackage{comment}
\usepackage{tikz}
\usetikzlibrary{arrows.meta, positioning, shapes.geometric, calc}
\usepackage[hidelinks, colorlinks=true, linkcolor=blue, urlcolor=blue, citecolor=blue]{hyperref}
\newcommand{\cmark}{\textcolor{green}{\ding{51}}} 
\newcommand{\xmark}{\textcolor{red}{\ding{55}}} 
\usepackage{threeparttable} 
\usepackage{colortbl} 
\definecolor{selected}{rgb}{0.88,1,1}
\definecolor{OliveGreen}{rgb}{0,0.6,0}
\usepackage{ragged2e}
\usepackage{amsmath}
\usepackage{bm}
\usepackage{xcolor}
\usepackage{algorithm}
\usepackage{algpseudocode}
\usepackage{graphicx}
\usepackage{textcomp}
\usepackage{subcaption}
\usepackage{multirow}
\justifying

\usepackage[table, svgnames, dvipsnames]{xcolor}
\usepackage{makecell, cellspace, caption}
\usepackage{multirow}

\begin{document}

\pagestyle{fancy}
\rhead{\includegraphics[width=2.5cm]{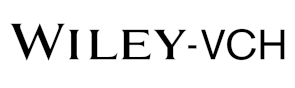}}

\title{\noindent Disentangling Aleatoric and Epistemic Uncertainty in \\ Physics-Informed Neural Networks. Application to\\ Insulation Material Degradation Prognostics}

\maketitle

\author{Ibai Ramirez$^1$}
\author{Jokin Alcibar$^1$}
\author{Joel Pino}
\author{Mikel Sanz}
\author{Jose I. Aizpurua\textbf{*}}


\dedication{}

\begin{affiliations}

\noindent I. Ramirez, J. Alcibar\\
Department of Electronics and Computer Science\\
Mondragon University\\
Arrasate-Mondragon, 20500, Spain\\
$^1$Equal Contribution \\

\noindent J. Pino\\
Research \& Development\\
CARTIF Technology Centre\\
Valladolid, 47151, Spain\\

\noindent M. Sanz \\
Department of Physical Chemistry \\
University of the Basque Country (EHU)\\
Basque Centre for Applied Mathematics (BCAM) \\
Leioa, 20018, Spain\\

\noindent Jose I. Aizpurua\\
Department of Computer Science and Artificial Intelligence \\
University of the Basque Country (EHU)\\
Donostia-San Sebastian, 20018, Spain\\
\textbf{*}Corresponding Author: joxe.aizpurua@ehu.eus

\end{affiliations}




\begin{abstract}

Physics-Informed Neural Networks (PINNs) provide a framework for integrating physical laws with data. However, their application to Prognostics and Health Management (PHM) remains constrained by the limited uncertainty quantification (UQ) capabilities. Most existing PINN-based prognostics approaches are deterministic or account only for epistemic uncertainty, limiting their suitability for risk-aware decision-making. This work introduces a heteroscedastic Bayesian Physics-Informed Neural Network (B-PINN) framework that jointly models epistemic and aleatoric uncertainty, yielding full predictive posteriors for spatiotemporal insulation material ageing estimation. The approach integrates Bayesian Neural Networks (BNNs) with physics-based residual enforcement and prior distributions, enabling probabilistic inference within a physics-informed learning architecture. The framework is evaluated on transformer insulation ageing application, validated with a finite-element thermal model and field measurements from a solar power plant, and benchmarked against deterministic PINNs, dropout-based PINNs (d-PINNs), and alternative B-PINN variants. Results show that the proposed B-PINN provides improved predictive accuracy and better-calibrated uncertainty estimates than competing approaches. A systematic sensitivity study further analyzes the impact of boundary-condition, initial-condition, and residual sampling strategies on accuracy, calibration, and generalization, and the influence of measurement noise on aleatoric uncertainty. Overall, the findings highlight the capability of Bayesian physics-informed learning to support uncertainty-aware prognostics and informed decision-making in transformer asset management by tracking aleatoric and epistemic sources of uncertainty.

\end{abstract}

\section{Introduction}

Prognostics and health management (PHM) is a health management paradigm that encompasses predictive methodologies aimed at improving the reliability and availability of engineering components and systems \cite{vachtsevanos2006_book}. PHM lies at the core of condition monitoring technologies, where engineering and physical principles, and monitored data are jointly used to develop anomaly detection,  diagnostics, prognostics, and maintenance planning applications. The main objective of failure prognostics models is the reliable prediction of the future degradation trajectory of an asset and the estimation of its remaining useful life (RUL)  \cite{Kordestani_21}.

Scientific Machine Learning (SciML) has emerged as a transformative paradigm for solving scientific and engineering problems by embedding physical structure directly into machine learning models \cite{karniadakis2021physicsinformed}. Representative developments in SciML include Physics-Informed Neural Networks (PINNs) \cite{Raissi_19}, Neural Operators \cite{NeuralOperators_23}, Physics-Informed Neural Operators (PINO) \cite{PINO_24}, Kolmogorov–Arnold Networks (KANs) \cite{KAN_ICLR}, and related approaches designed to learn solutions of governing equations or physical dynamics from data \cite{toscano2025}. By incorporating partial differential equations (PDEs) and mechanistic constraints during training, these models typically achieve improved sample efficiency and enhanced generalization compared with purely data-driven methods.

Within the broader PHM literature, hybrid prognostics approaches have historically combined physics-of-failure models with machine learning methods for RUL estimation, relying on sound engineering and physical principles \cite{aizpurua2015towards, Guo_20, Zio_22}. Such hybrid approaches generally follow either a sequential architecture, in which physics-based model outputs inform a machine learning model \cite{Chao22, daigle2015model}, or a parallel architecture, where physics-based and machine learning predictions are fused \cite{Aizpurua_23,Alcibar_2025}. Hybrid prognostics approaches share conceptual similarities with SciML, but they usually lack a unified learning framework in which physics and data jointly shape the solution.

In prognostics, the RUL is defined as the temporal distance between the prediction time instant, denoted by $t_p$, and the end of the useful life of the asset, referred to as the end-of-life (EOL). This quantity is commonly estimated by forecasting the future ageing evolution of the system and identifying the expected failure threshold crossing time. The ageing evolution from $t_p$ to EOL is inherently stochastic and affected by multiple sources of uncertainty, including sensor noise, operational variability, and model uncertainty. Consequently, uncertainty quantification (UQ) is essential for reliable RUL predictions. 

The application of PINNs to prognostics has recently been explored in domains such as battery degradation modelling \cite{wang2024physics} and transformer ageing based on heat diffusion physics \cite{Ramirez_25}. However, to the best of the authors' knowledge, the majority of existing PINN-based prognostics models remain fully deterministic, despite the critical importance of uncertainty quantification for informed decision-making \cite{sankararaman2015significance, Salinas_25}. This limitation is particularly restrictive in realistic PHM scenarios, where data are often sparse, noisy, and heterogeneous, and where multiple sources of uncertainty, such as model misspecification, parameter variability, and measurement noise, must be explicitly accounted for.

Within the SciML community, several uncertainty-aware PINN formulations have been proposed. Bayesian PINNs (B-PINNs) replace deterministic Neural Networks (NNs) with Bayesian Neural Networks (BNNs), enabling the estimation of epistemic uncertainty \cite{Linka2022}. Alternative approaches include Gaussian process–augmented PINNs \cite{Li_2024}, Hamiltonian Monte Carlo–based epistemic PINNs (EPINNs) \cite{Epinet2025}, conformalized PINNs (C-PINNs) providing distribution-free uncertainty guarantees \cite{podina2024conformalized}, as well as more computationally tractable methods such as dropout-PINNs (d-PINNs) \cite{Zhang_19} and deep ensemble PINNs \cite{jiang2023practical, zou2025learning}. Despite these methodological advances, most studies are restricted to synthetic PDE benchmarks, and their applicability to real-world prognostics problems remains largely unexplored.

More importantly, existing uncertainty-aware PINN approaches rarely provide a comprehensive treatment of total predictive uncertainty by explicitly disentangling epistemic and aleatoric components \cite{hullermeier2021aleatoric}. Epistemic uncertainty reflects uncertainty in the model structure and parameters due to limited data, whereas aleatoric uncertainty arises from measurement noise and inherent system variability. In PHM applications, the separation and propagation of both uncertainty sources are critical for robust risk assessment and maintenance decision-making \cite{sankararaman2015significance, Salinas_25}.

Table~\ref{tab:uq_comparison} summarizes representative SciML and hybrid PHM approaches, highlighting whether they model aleatoric and/or epistemic uncertainty, explicitly disentangle uncertainty sources, adhere to SciML principles, and address PHM applications.

\begin{table}[!htb]
	\centering
	\setlength{\tabcolsep}{3.500pt}
	\caption{Comparison of SciML and hybrid PHM methods, including relevant features for this work.}
	\label{tab:uq_comparison}
	\begin{tabular}{lcccccc}
		\toprule
		\textbf{Method} & \textbf{Aleatoric} & \textbf{Epistemic} & \textbf{Disentanglement} & \textbf{SciML} & \textbf{PHM} & \textbf{Method} \\
		\midrule
		\cite{Juan_2023}         & \xmark & \cmark & \xmark & \xmark & Material & BNN \\
		\cite{nascimento2023framework} & \cmark & \cmark & \cmark & \xmark & Battery & RNN \\
		\cite{wang2024physics}         & \xmark & \xmark & \xmark & \cmark & Battery & PINN \\
		\cite{Ramirez_25} & \xmark & \xmark & \xmark & \cmark &  Transformer & PINN  \\
		\cite{Linka2022} & \xmark & \cmark & \xmark & \cmark & \xmark & B-PINN\\
		\cite{podina2024conformalized}  & \cmark & \xmark & \xmark & \cmark & \xmark & C-PINN\\
		\cite{Epinet2025}         & \xmark & \cmark & \xmark & \cmark & \xmark & EPINN \\
		\cite{Zhang_19}  & \cmark & \cmark & \xmark & \cmark & \xmark &  d-PINN\\
		\cite{zou2025learning}  & \cmark & \cmark & \xmark & \cmark & \xmark &  EnsPINN\\
		\rowcolor{Gainsboro!60} 
        \textbf{This Work}     & \cmark & \cmark & \cmark & \cmark & Transformer &  B-PINN\\
		\bottomrule
	\end{tabular}
\end{table}


In this context, it can be observed that PINN-based prognostics models have predominantly relied on deterministic formulations. While uncertainty-aware PINNs have been proposed within the SciML community, to the best of the authors' knowledge, they (i) do not explicitly disentangle aleatoric and epistemic sources of uncertainty, and (ii) have largely been evaluated on controlled synthetic benchmarks rather than realistic PHM applications.


Motivated by the need for uncertainty-aware SciML methods for reliable prognostics, the main contribution of this work is the development of a total UQ framework for PINNs that jointly captures aleatoric and epistemic uncertainty in prognostics applications. The proposed methodology is based on B-PINNs, which integrate PDE-based ageing physics with Bayesian inference over Neural Network (NN) parameters. Extending previous work that mainly focused on epistemic UQ \cite{BPINN_PHM25}, a novel B-PINN architecture is introduced that enables explicit modelling and separation of epistemic and aleatoric sources of uncertainty.

This work demonstrates that incorporating epistemic and heteroscedastic aleatoric uncertainty within B‑PINNs yields more reliable and interpretable probabilistic predictions than B‑PINNs that model epistemic and homoscedastic uncertainty, as well as deterministic and dropout‑based PINN alternatives. The proposed methodology improves predictive accuracy, enhances uncertainty calibration, and increases robustness under noisy and sparse data,  which are essential properties for enabling trustworthy prognostics predictions in critical power assets.

This research further contributes with a systematic analysis of the influence of initial-condition, boundary-condition, and residual sampling strategies on generalization and uncertainty calibration, offering new insights into training stability and robustness. Moreover, the influence of measurement noise errors on aleatoric uncertainty is also analysed showing that aleatoric uncertainty increases in agreement with data uncertainty, while epistemic uncertainty remains stable. The proposed approach is validated with a real-world transformer insulation ageing case study, supported by finite element modelling and operational data from a floating solar power plant. The validated results are supported by a comprehensive benchmarking, which compares B-PINNs with d-PINNs in two different total UQ configurations: (i) epistemic and heteroscedastic aleatoric UQ and (ii) epistemic and homoscedastic aleatoric UQ, along with vanilla deterministic PINNs baselines.

The remainder of this article is organized as follows. Section~\ref{sec:Methodology} presents the proposed methodology. Section~\ref{s:Experimental} describes the experimental setup, results, and sensitivity analyses. Section~\ref{s:Conclusions} concludes the paper. Appendix~\ref{s:Trafo} details the transformer thermal and ageing models and Appendix~\ref{s:benchmarking_metrics} defines the probabilistic evaluation metrics and benchmarking models.

\section{Methods}
\label{sec:Methodology}

\begin{figure}[!htb]
	\centering
	\includegraphics[width=0.95\linewidth]{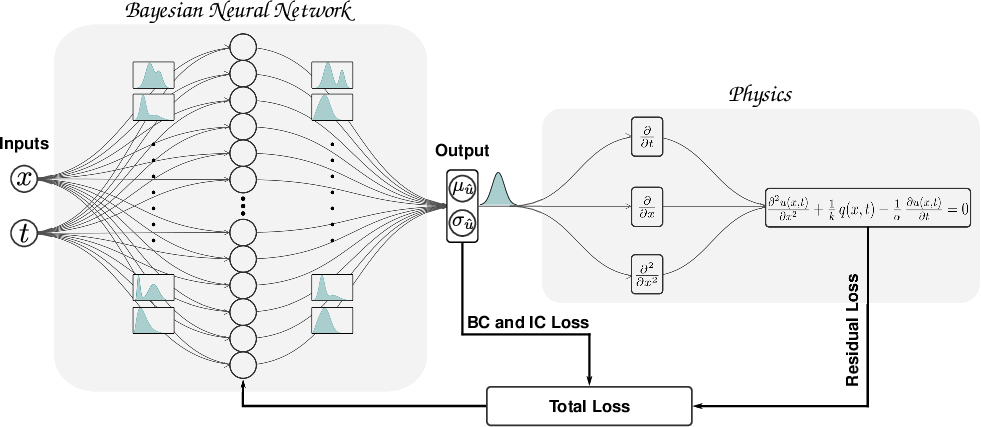}
	\caption{Proposed B-PINN approach for the probabilistic spatiotemporal transformer thermal model, including spatiotemporal inputs, $(x, t)$, a BNN with probability distribution function over its parameters, two output variables over the solution ($u$) of the PDE under study ($\mu_{\hat{u}}$, $\sigma_{\hat{u}}^2$), the physics residual loss, and initial and boundary condition losses integrated in the total loss, which is used to update BNN parameter distributions.}
	\label{fig:BPINN_Framework_General}
\end{figure}

Figure~\ref{fig:BPINN_Framework_General} shows the proposed methodology for the total UQ of Bayesian-PINN models, including epistemic and (heteroscedastic) aleatoric sources of uncertainty.

BNNs are used to integrate Bayesian inference into PINNs, enabling principled uncertainty quantification. BNNs capture the epistemic uncertainty of the B-PINN model by modelling the uncertainty associated with the NN parameters. For the quantification of aleatoric uncertainty, a modification of the classical B-PINN model architecture presented in \cite{BPINN_PHM25} is proposed. Assuming that the uncertainty can be captured with an equivalent Gaussian distribution, the proposed B-PINN model outputs two variables, the mean value of the solution $\mu_{\hat{u}}$ and the variance of the solution $\sigma_{\hat{u}}^2$. The adoption of these two outputs requires modifying the loss function so that the B-PINN learning process is directed by the underlying PDE, regularized by Bayesian principles, and integrates aleatoric and epistemic uncertainties (detailed in Subsection~\ref{ss:BPINN}).

The proposed approach is tested on the electrical transformer insulation material ageing estimation. The insulation ageing estimation approach is based on a spatiotemporal thermal stress model, which drives the insulation ageing model, modeled with an empirical ageing estimation approach. 

\begin{figure}[!htb]
	\centering
	\includegraphics[width=.7\columnwidth]{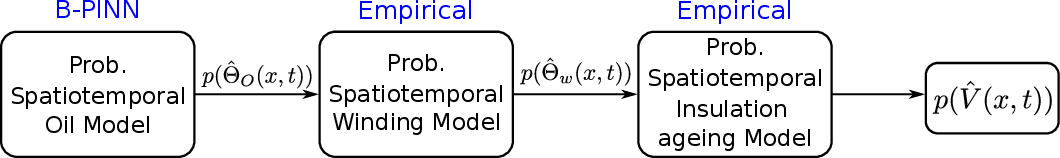}
	\caption{Overall framework for probabilistic transformer insulation ageing estimation. The Bayesian PINN-based model estimates the spatiotemporal oil temperature of the insulation, $\hat{\Theta}_o(x,t)$, which is coupled with an empirical model, which calculates the transformer winding temperature, $\hat{\Theta}_w(x,t)$, and finally, the estimated thermal stress is used to calculate the insulation ageing model $p(\hat{V}(x,t))$, through an empirical approach.}
	\label{fig:Framework_General}
\end{figure}

The thermal stress estimation is done in two steps: (i) the insulation oil temperature estimation, $\hat{\Theta}_o(x,t)$, which is calculated using a 1D heat diffusion PDE with Dirichlet boundary conditions, implemented using the proposed B-PINN approach. (ii) The oil temperature is used to calculate the spatiotemporal winding temperature, $\hat{\Theta}_w(x,t)$, through an empirical winding temperature estimation model. Finally, the insulation ageing estimate, $\hat{V}(x,t)$, is obtained by using the winding temperature, which defines the ageing rate and it is directly used to estimate the insulation RUL \cite{Aizpurua_22}. Modelling epistemic and aleatoric sources of uncertainty enables the quantification of the probabilistic estimates of oil temperature, $p(\hat{\Theta}_o(x,t))$, winding temperature, $p(\hat{\Theta}_w(x,t))$, and ageing estimates, $p(\hat{V}(x,t))$. Figure~\ref{fig:Framework_General} shows the insulation ageing estimation process.  

The main focus of this section is on the B-PINN method for total uncertainty quantification. Detailed insulation thermal modelling process, along with insulation ageing estimation modelling equations are provided in Appendix~\ref{s:Trafo}. Accordingly, Subsection~\ref{ss:PINN_Basics} introduces PINN basics, Subsection~\ref{ss:BNN_Basics} defines BNNs, Subsection~\ref{ss:BPINN} describes the B-PINN approach, and finally Subsection \ref{ss:UQ} defines the total uncertainty quantification process.

\subsection{PINN basics}
\label{ss:PINN_Basics}

PINNs were introduced with the goal of encoding physics models based on PDEs in machine learning (ML) models \cite{Raisi_19}, taking advantage of the ability of NNs  to act as universal approximators \cite{wright2022deep}. A general PDE can be written as \cite{Agnostopoulos_RBA}:

\begin{equation}
	\label{eq:PDE_generic}
	d\{u(x,t)\}=f(x,t)
\end{equation}

\noindent where $x,t \in \mathbb{R}$ denote position and time, $u(x,t)$ is the unknown solution, $d$ is the differential operator, and $f(x,t)$ is a forcing function introducing external influences.

PINNs consist of the NN part, in which the inputs define temporal ($t \in \mathbb{R}$) and spatial coordinates ($x \in \mathbb{R}$ for one-dimensional cases) for the initial conditions (IC) and boundary conditions (BC). The NN output is an approximated PDE solution at the space and time coordinates, denoted {$\hat{u}(x,t)$}. This is calculated through the iterative application of weights (\bm{$w$}), biases (\bm{$b$}), and non-linear activation functions ($\sigma$) over the input. Namely, the inputs are connected through neurons, where they are multiplied with the weights and summed with the bias term. Finally, the weighted sum is passed through an activation function ($\sigma$). Subsequently, the outcome {$\hat{u}(x,t)$} is post-processed via automatic differentiation to compute the derivatives in space and time at certain collocation points (CP), generated via random sampling in the interior of the domain. The PINN is trained at these CPs by minimizing the residuals of the underlying PDE. 

The loss function, $\mathcal{L}(\bm{\theta},\lambda_0,\lambda_b,\lambda_r)$, incorporates the prediction error of the NN at IC and BC, and the residual of the PDE estimated via automatic differentiation at CP:

\begin{equation}
	\label{eq:Loss_PINN_generic}
	\mathcal{L}(\bm{\theta},\lambda_0,\lambda_b,\lambda_r)=\mathcal{L}_0(\bm{\theta},\lambda_0)+\mathcal{L}_b(\bm{\theta},\lambda_b)+\mathcal{L}_r(\bm{\theta},\lambda_r)
\end{equation}

\noindent where {$\bm{\theta}$\hspace{0.5mm}=\hspace{0.5mm}$\{\bm{w,b}\}$} are the weights and bias terms of the NN, and $\mathcal{L}_0(\bm{\theta},\lambda_0),\mathcal{L}_b(\bm{\theta},\lambda_b)$, and $\mathcal{L}_r(\bm{\theta},\lambda_r)$ are, respectively, the loss terms corresponding to IC, BC, and the residual of the PDE with their corresponding weights, $\lambda_0$, $\lambda_b$, and $\lambda_r$, defined as follows:

\begin{equation}
	\label{eq:Loss_IC}
	\mathcal{L}_0(\bm{\theta},\lambda_0)=\lambda_0\frac{1}{N_{0}}\sum_{i=1}^{N_{0}}|\hat{u}(x_i,0)-u(x_i,0)|^2
\end{equation}

\begin{equation}
	\label{eq:Loss_BC}
	\mathcal{L}_b(\bm{\theta},\lambda_b)=\lambda_b\frac{1}{N_{b}}\sum_{i=1}^{N_{b}}|\hat{u}(x_i,t_i)-u(x_i,t_i)|^2
\end{equation}

\begin{equation}
	\label{eq:Loss_r}
	\mathcal{L}_r(\bm{\theta},\lambda_r)=\lambda_r\frac{1}{N_r}\sum_{i=1}^{N_r}|r(x_i,t_i)|^2
\end{equation}

\noindent where $N_0,N_b$, and $N_r$ are, respectively, the number of IC, BC, and residual points, while $u(x_i,t_i)$ and $r(x_i,t_i)$ denote the known solution and the residual of PDE, for each training point $i$ defined at the coordinates $(x_i,t_i)$. According to the PDE defined in Eq.~(\ref{eq:PDE_generic}), the residual {$r(x,t)$} is defined as follows:

\begin{equation}
	\label{eq:Residual_generico}
	r(x,t)=d\{u(x,t)\}-f(x,t)
\end{equation}

Minimizing the loss function in Eq.~(\ref{eq:Loss_PINN_generic}) using a suitable optimization algorithm provides an optimal set of NN parameters $\bm{\theta}=\{\bm{w,b}\}$. That is, approximating the PDE becomes equivalent to finding $\bm{\theta}$ values that minimize the loss with a predefined accuracy. Overall, the key training parameters include: number of neurons and number of layers, number of CP, activation function, and the optimizer. Finding the correct solution requires knowing IC and BC. Additionally, random locations $(x_i,t_i)$, named CP, are used to evaluate the residual loss in Eq.~(\ref{eq:Loss_r}).

\subsection{Bayesian Neural Networks}
\label{ss:BNN_Basics}

BNNs aim to estimate the posterior parameter distribution $P(\bm{\theta}|\mathcal{D})$, from a training dataset $\mathcal{D}=\{\bm{x}^{(i)},\bm{y}^{(i)}\}$, of a set of NN parameters $\bm{\theta}$:

\begin{equation}
	P(\bm{\theta}|\mathcal{D})=\frac{P(\mathcal{D}|\bm{\theta})P(\bm{\theta})}{P(\mathcal{D})}
	\label{eq:Bayes}
\end{equation}

\vspace{-1 mm}

\noindent where $P(\mathcal{D}|\bm{\theta})$ is the likelihood, $P(\bm{\theta})$ is the prior, \textit{i.e.} prior knowledge of NN parameters expressed as a PDF over $\bm{\theta}$, and $P(\mathcal{D})$ is the marginal likelihood.

The likelihood is often estimated from the individual product of pointwise estimated likelihoods $p_i(\mathcal{D}|\bm{\theta})$ based on the normal distribution $N(\mu,\sigma)$ defined as follows:

\begin{equation}
	P(\mathcal{D}|\bm{\theta})=\prod_{i=0}^{N}  p(y^{(i)}|x^{(i)},\bm{\theta})
\end{equation}

\begin{equation}
\label{eq:Gaussian}
p(y^{(i)} \mid x^{(i)}, \bm{\theta})
= \frac{1}{\sqrt{2\pi}\sigma}
\exp\left(-\frac{\| y^{(i)} - \hat{u}(x^{(i)};\bm{\theta}) \|^2}{2\sigma^2}\right)
\end{equation}

The selection of a good prior distribution, $P(\bm{\theta})$, for BNNs is challenging \cite{Fortuin_21}. The most common choice is to use a non-informative prior, which follows a Gaussian distribution with mean zero and unit variance, $\mathcal{N}(0,1)$, \textit{i.e.} isotropic Gaussian prior. In addition to the isotropic Gaussian prior, there are alternative non-informative priors that can be used for an improved posterior prediction in terms of accuracy and uncertainty. Without loss of generality, based on previous experience testing different priors \cite{BPINN_PHM25}, this research implements the Laplace prior, defined as follows:

\begin{equation}
	\label{eq:L_prior}
	P(\bm{\theta})_{Laplace} = \prod_{i=1}^{d} \frac{\lambda}{2} \exp\left(-\lambda |\theta_i|\right)
\end{equation}

\noindent where $\lambda > 0$ is the scale parameter, and $d$ is the dimensionality of $\bm{\theta}$. The Laplace prior also promotes sparsity due to its sharp peak at zero and heavier tails compared to the non-informative Gaussian prior \cite{Fortuin_21}.

The analytical posterior in BNNs, $p(\bm{\theta}|\mathcal{D})$, is intractable, and therefore approximation methods are required. In this work, Variational Inference (VI) is used to approximate the true posterior with a variational distribution $q(\bm{\theta}|\bm{\phi})$ of known functional form by minimizing the Kullback–Leibler (KL) divergence between them. Since the true posterior cannot be computed explicitly, this optimization is performed indirectly by maximizing the Evidence Lower Bound (ELBO). In practice, as learning is formulated as a minimization problem, ELBO maximization is equivalenty implemented by minimizing its negative, commonly referred to the variational free energy.  The resulting optimization objective is defined as follows:

\begin{equation}
\label{eq:KL}
	\mathcal{L}(\mathcal{D}, \bm{\phi}) = \text{KL}[q(\bm{\theta} \mid \bm{\phi}) \parallel p(\bm{\theta})] - \mathbb{E}_{q(\bm{\theta} \mid \bm{\phi})}[\log p(\mathcal{D} \mid \bm{\theta})]
\end{equation}

The first term is the KL divergence between the variational distribution $q(\bm{\theta}|\bm{\phi})$ and the prior $p(\bm{\theta})$ (known as the complexity cost). The second term is the expected value of the likelihood with respect to the variational distribution (known as the likelihood cost). The cost function can be also written as:

\begin{equation}
	\begin{split}
		\label{eq:VI_log}
		\mathcal{L}(\mathcal{D}, \bm{\phi}) = 
		\mathbb{E}_{q(\bm{\theta}|\bm{\phi})} \left[ \log q(\bm{\theta}|\bm{\phi}) \right] 
		- \mathbb{E}_{q(\bm{\theta}|\bm{\phi})} \left[ \log p(\bm{\theta}) \right] 
		- \mathbb{E}_{q(\bm{\theta}|\bm{\phi})} \left[ \log p(\mathcal{D}|\bm{\theta}) \right]
	\end{split}
\end{equation}

It can be observed from Eq.~(\ref{eq:VI_log}) that loss terms are expectations with respect to the variational distribution $q(\bm{\theta}|\bm{\phi})$. Therefore, the cost function can be approximated by Monte Carlo (MC) sampling, drawing $K$ samples $\bm{\theta}^{(k)}$ from $q(\bm{\theta}|\bm{\phi})$:

\begin{equation}
	\mathcal{L}(\mathcal{D},\!\bm{\phi})\! \approx \!
	\frac{1}{K}\! \sum_{k=1}^{K}\!\big[
	\log q(\bm{\theta}^{(k)}|\bm{\phi}) 
	\!-\! \log p(\bm{\theta}^{(k)}) 
	\!-\! \log p(\mathcal{D}|\bm{\theta}^{(k)}) \big]
	\label{eq:MC_ELBO}
\end{equation}

Throughout this paper, the variational posterior is assumed to follow a Gaussian distribution $\bm{\phi}=(\bm{\mu},\bm{\sigma})$, where $\bm{\mu}$ is the mean vector of the distribution and $\bm{\sigma}$ is the standard deviation vector.

\subsection{Bayesian Physics Informed Neural Networks}
\label{ss:BPINN}

Bayesian PINNs extend standard PINNs by replacing the deterministic NN with a Bayesian NN. This enables the estimation of epistemic uncertainty, which reflects the model’s lack of complete knowledge~\cite{kendall2017}. In practice, epistemic uncertainty is quantified by placing probability density functions (PDFs) over the BNN parameters.

The main outcome of interest in the B-PINN approach is the posterior distribution. This is calculated following the Bayes rule [cf. Eq. ~(\ref{eq:Bayes})]. Namely, the prior distribution is assumed to be given based on \textit{a priori} knowledge. The likelihood, $P(\mathcal{D}|\bm{\theta})$, is estimated for the initial point, boundary condition points, and residual samples assuming a Gaussian distribution [cf. Eq.~(\ref{eq:Gaussian})], defined as follows:

\begin{align}
	\begin{split}
		& p(u_0 | x_0, \bm{\theta}^{(k)}) = \prod_{i=1}^{N_0} \frac{1}{\sqrt{2\pi} \sigma_0}\exp\left(-\frac{\| u_0-\hat{u}(x^{(i)}_0, 0; \bm{\theta}^{(k)}) \|^2}{2 \sigma_0^2}\right)\\
		& p(u_{bc} | x_{bc}, t_{bc}, \bm{\theta}^{(k)}) =\prod_{i=1}^{N_{bc}} \frac{1}{\sqrt{2\pi} \sigma_{bc}}\exp\left(-\frac{\| u_{bc} -\hat{u}(x^{(i)}_{bc}, t^{(i)}_{bc}; \bm{\theta}^{(k)}) \|^2}{2 \sigma_{bc}^2}\right)\\
		& p(r(x_f, t_f; \bm{\theta}^{(k)})) = \prod_{i=1}^{N_{f}} \frac{1}{\sqrt{2\pi} \sigma_{f}}\exp\left(-\frac{\| r(x^{(i)}_f, t^{(i)}_f; \bm{\theta}^{(k)}) \|^2}{2 \sigma_{f}^2}\right)
	\end{split}
	\label{eq:L_BPINN}
\end{align}

\noindent where $\sigma_0$, $\sigma_{bc}$, and $\sigma_f$ denote the standard deviation of initial, boundary, and residual points. 

For computational tractability and efficiency, the log-likelihood terms are considered by taking the logarithm of Eq. (\ref{eq:L_BPINN}). Subsequently, the ELBO loss defined for a generic BNN [cf. Eq.~(\ref{eq:MC_ELBO})], is adapted for B-PINN posterior inference:

\begin{align}
\begin{split}
		\mathcal{L}^{(k)} = 
		&\log q(\bm{\theta}^{(k)}|\bm{w}) 
		- \log p(\bm{\theta}^{(k)}) 
		- \lambda_0\log p(u_0 | x_0, \bm{\theta}^{(k)}) - \\
		& \lambda_b\log p(u_{bc} | x_{bc}, t_{bc}, \bm{\theta}^{(k)})
		-\lambda_r \log p(r | x_f, t_f, \bm{\theta}^{(k)})
	\label{eq:ELBO_BPINN}
\end{split}
\end{align}

This process results in the approximation of the variational posterior distribution  $q(\bm{\theta}|\bm{w})$. The training process of the B-PINN approach is shown in Appendix~\ref{s:BPINN_PHM}, Algorithm~\ref{alg:Bayesian_PINN}.

It is worth highlighting that the variance terms in Eq.~(\ref{eq:L_BPINN}), if kept constant, they model constant aleatoric uncertainty, known as homoscedastic uncertainty, \textit{i.e.} constant variance. The limitation of homoscedastic uncertainty is that the uncertainty is fixed by the user and held constant throughout the analysis. If this assumption can be relaxed (as proposed in this work), automatically learning aleatoric uncertainty from data presents a data-adaptive uncertainty modelling strategy, \textit{i.e.} heteroscedastic uncertainty.

\subsection{Uncertainty Quantification}
\label{ss:UQ}

The modified B-PINN model provides predictive distributions by (i) placing a posterior over the network parameters, which induces epistemic uncertainty, and (ii) producing an input-dependent predictive variance, capturing heteroscedastic aleatoric uncertainty.

\subsubsection*{Aleatoric uncertainty}
\label{ss:Aleatoric}

Aleatoric uncertainty captures the inherent randomness in the data-generating process. It represents irreducible noise (e.g., measurement noise) that cannot be mitigated by collecting more data~\cite{kendall2017}. Aleatoric uncertainty can be homoscedastic, \textit{i.e.} constant across the domain, or heteroscedastic, where noise varies with the input.

Widely adopted prognostics methods that assume homoscedastic uncertainty include Particle Filtering approaches~\cite{Marcos_PF_2009}. In contrast, real transformer insulation degradation data exhibit heteroscedasticity driven by load fluctuations, thermal inertia, and operational variability.

In heteroscedastic regression, the modified B-PINN outputs both the predictive mean $\hat{u}(x,t;\bm{\theta})$ and an input-dependent variance \(\sigma^2(x,t;\bm{\theta})\). Under a Gaussian likelihood, the log-likelihood per ground truth sample $u$, corresponding to the spatiotemporal coordinates $(x, t)$ is defined as follows:

\begin{equation}
\label{eq:loglik_hetero}
\log p\big(u \mid x,t;\bm{\theta}\big)
= -\tfrac{1}{2}\log\big(2\pi\,\sigma^2(x,t;\bm{\theta})\big)
- \frac{\big(u - \hat{u}(x,t;\bm{\theta})\big)^2}{2\,\sigma^2(x,t;\bm{\theta})}.
\end{equation}

During Monte Carlo inference (cf. Algorithm~\ref{alg:Bayesian_PINN}), the aleatoric uncertainty is estimated by averaging the predictive variance across the $K$ posterior samples:

\begin{equation}
\label{eq:aleatoric_mc}
\widehat{\mathrm{AU}}(x,t) = \frac{1}{K}\sum_{k=1}^K \sigma^2(x,t;\bm{\theta}_k)
\end{equation}

\noindent
where $\sigma^2(x,t;\bm{\theta}_k)$ is the network’s variance output for parameter sample $\bm{\theta}_k$.

\subsubsection*{Epistemic uncertainty}
\label{ss:EpistemicUncertainty}

Epistemic uncertainty represents uncertainty in the model parameters due to limited data or incomplete knowledge of the underlying system. Unlike aleatoric uncertainty, epistemic uncertainty is reducible with more informative data or stronger priors~\cite{hullermeier2021}.

Bayesian PINNs quantify epistemic uncertainty by sampling from the variational posterior $q(\theta)$. With $K$ total samples, epistemic uncertainty is estimated as the predictive variance across the Monte Carlo forward passes:

\begin{equation}
\label{eq:epistemic_mc}
    \widehat{\mathrm{EU}}(x,t) = \frac{1}{K}\sum_{k=1}^{K} {\hat{u}}(x,t;\bm{\theta_k})^2 - \left(\frac{1}{K}\sum_{k=1}^{K} {\hat{u}}(x,t;\bm{\theta_k}) \right)^2
\end{equation}

\noindent
where ${\hat{u}}(x,t;\bm{\theta_k})$ is the model prediction using parameter sample $\bm{\theta}_k$.

\subsubsection*{Total predictive uncertainty}
\label{ss:TotalUncertainty}

Using variance as the uncertainty metric, the total predictive uncertainty can be decomposed using the law of total variance~\cite{Depeweg2018}:

\begin{equation}
\sigma^2(u|x) = \underbrace{\sigma^2_\theta[E_{u|x,\theta}(u|x, \theta)]}_{\text{epistemic}} + \underbrace{E_\theta[\sigma^2_{u|x,\theta}(u|x, \theta)]}_{\text{aleatoric}}
\label{eq:predictive_varicance}
\end{equation}

\noindent where $\sigma^2_\theta[E_{u|x,\theta}(u|x, \theta)]$ represents the epistemic uncertainty and $E_\theta[\sigma^2_{u|x,\theta}(u|x, \theta)]$ represents the aleatoric uncertainty.

\subsubsection*{B-PINNs Training with Total Uncertainty Quantification}

In the B-PINN setting without aleatoric heteroscedastic uncertainty, the loss in Eq.~(\ref{eq:KL}) is minimized to estimate the variational posterior. In order to incorporate the heteroscedastic uncertainty, the Gaussian heteroscedastic log-likelihood in Eq.~(\ref{eq:loglik_hetero}) is adopted, where the contribution of a spatio-temporal data point $(x,t)$ to the expected negative log-likelihood (NLL) is defined as follows:

\begin{equation}
\label{eq:NLL2}
\mathrm{NLL}_{\text{hetero}}(x,t;\bm{\theta})
= \tfrac{1}{2}\log\big(2\pi\sigma_{\hat{u}}^2(x,t;\bm{\theta})\big) + \frac{\big(u - \mu_{\hat{u}}(x,t;\bm{\theta})\big)^2}{2\sigma_{\hat{u}}^2(x,t;\bm{\theta})}
\end{equation}

\noindent where $u$ is the solution or ground truth, $\mu_{\hat{u}}(x,t;\bm{\theta})$ is the mean value of the solution, and $\sigma_{\hat{u}}^2(x,t;\bm{\theta})$ is the variance of the solution calculated from the proposed B-PINN architecture (cf. Figure~\ref{fig:BPINN_Framework_General}).

When VI is used to approximate the posterior distribution, the NLL terms associated with the initial conditions, boundary conditions, and residuals are incorporated into the ELBO. Each contribution is scaled by its corresponding weighting coefficient $\lambda$. In practice, the optimization is formulated as a minimization problem [cf. Eq.~(\ref{eq:KL})]. Consequently, the per-sample Monte Carlo training objective corresponds to minimizing the variational free energy, obtained by combining the KL divergence term with the weighted NLL components. The resulting loss function used during training is given by:

\begin{align}
\label{eq:elbo_hetero_final}
\mathcal{L}^{(k)} \;=\; &\log q(\bm{\theta}^{(k)} \mid \phi) - \log p(\bm{\theta}^{(k)}) \\
&+ \lambda_0\,\mathrm{NLL}_{\text{hetero}}^{(0)}(\bm{\theta}^{(k)})
+ \lambda_b\,\mathrm{NLL}_{\text{hetero}}^{(\mathrm{bc})}(\bm{\theta}^{(k)})
+ \lambda_r\,\mathrm{NLL}_{\text{hetero}}^{(r)}(\bm{\theta}^{(k)}) \nonumber
\end{align}

\noindent where each $\mathrm{NLL}_{\text{hetero}}^{(.)}(\bm{\theta}^{(k)})$ term is the sum over the corresponding initial condition, boundary condition and residual points of Eq.~(\ref{eq:NLL2}).

Algorithm~\ref{alg:total_uncertainty} summarizes the total predictive uncertainty inference process for B-PINNs.

\begin{algorithm}[ht]
    \caption{Total predictive uncertainty inference for Bayesian PINNs.}
    \label{alg:total_uncertainty}
    \begin{algorithmic}[1]
    \State \textbf{Input:} B-PINN model $f(x,t;\bm{\theta})$, test location $(x,t)$, posterior samples $\{\theta_k\}_{k=1}^K$
    \For{$k=1$ to $K$}
        \State $[\mu_{\hat{u}}, \sigma_{\hat{u}}^2] = f(x,t; \bm{\theta})$
        \Comment{Predictive mean and aleatoric variance}
        \State Store $\mu_{\hat{u}}$ in vector $\vec{u}$
        \State Store $\sigma_{\hat{u}}^2$ in vector $\vec{\sigma}^2$
    \EndFor
    \State $\widehat{u}(x,t) \leftarrow \frac{1}{K}\sum_{k}\vec{u}_k$
    \State $\widehat{\mathrm{EU}}(x,t) \leftarrow \frac{1}{K}\sum_{k}\vec{u}_k^2 - \widehat{u}(x,t)^2$
    \State $\widehat{\mathrm{AU}}(x,t) \leftarrow \frac{1}{K}\sum_{k} \vec{\sigma}_k$
    \State $\sigma_{\mathrm{TU}}(x,t) = \widehat{\mathrm{EU}}(x,t) + \widehat{\mathrm{AU}}(x,t)$
    \State \textbf{return} $\widehat{u}(x,t),\; \sigma_{\mathrm{TU}}(x,t)$
    \end{algorithmic}
\end{algorithm}

\section{Experiments}
\label{s:Experimental}

The case study examines the lifetime of a distribution transformer installed in a floating solar power plant in Spain \cite{Aizpurua_23}. Details of the distribution transformer are summarized in Appendix \ref{ss:TrafoDetails}.

\begin{figure}[!htb]
	\centering
	\includegraphics[width=0.5\columnwidth]{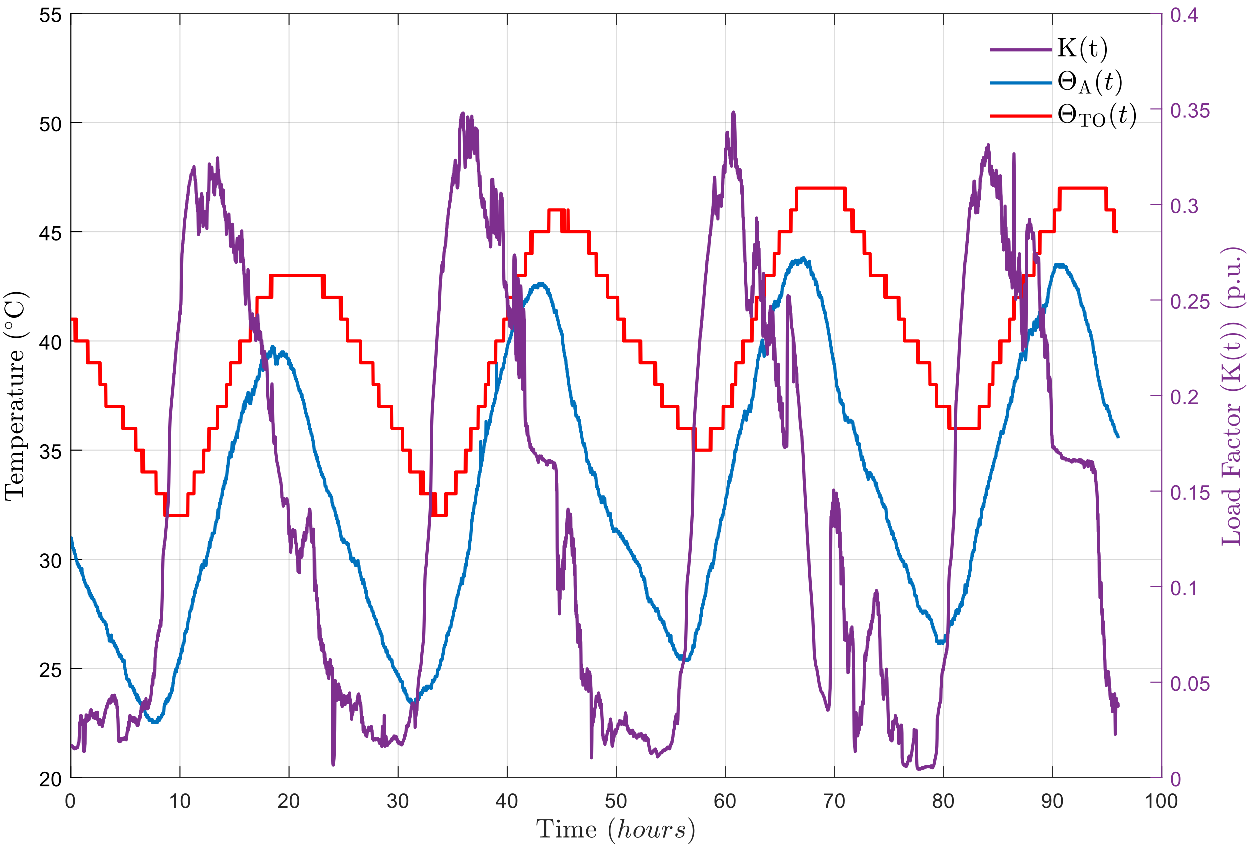}
	\caption{Distribution transformer load ($K$), ambient temperature ($\Theta_A$), and top-oil temperature ($\Theta_{TO}$) over four days of operation with one-minute resolution, operated at a floating photovoltaic substation.}
	\label{fig:AvailableTimeSeries}
\end{figure}

The dataset was preprocessed by removing variables without measurements, duplicate entries, and incorrect sensor readings. Missing values were imputed with average values. Due to the large thermal time constant of the transformer oil (see Table~\ref{table:CaseStudy_trafo}), temperature dynamics evolve slowly. Therefore, mean-value imputation strategy has negligible impact \cite{Aizpurua_23,Ramirez_25, BPINN_PHM25}. Figure~\ref{fig:AvailableTimeSeries} shows the available minutely sampled time series for ambient temperature, oil temperature, and load with a total of 5760 samples over 4 days of operation. Oil and winding temperature profiles analysed in this study are based on this minute-level sampling.

\begin{figure}[!htb]
	\centering
	\includegraphics[width=.5\columnwidth]{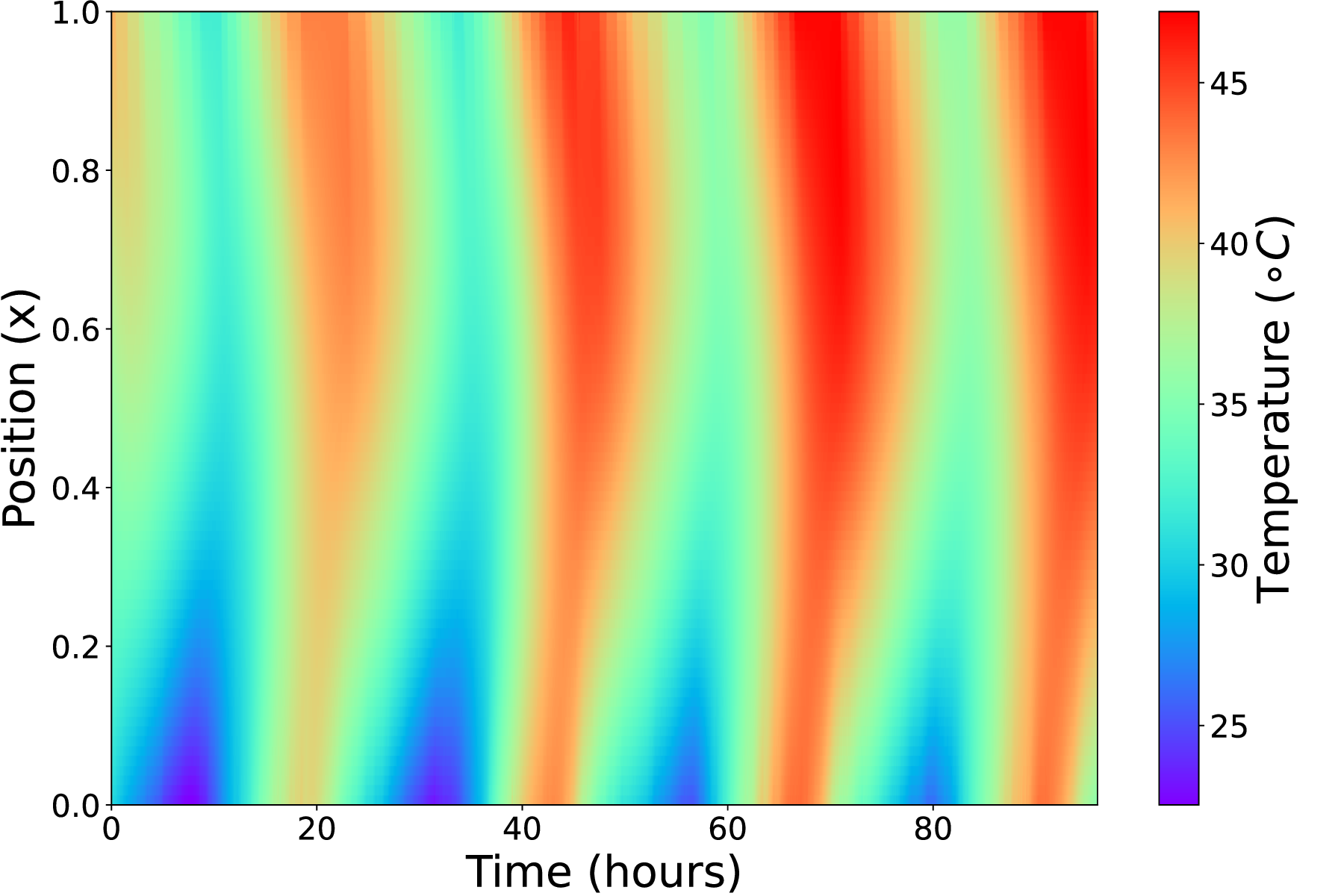}
	\caption{Spatiotemporal transformer oil temperature estimation obtained through a finite element method using the data in Figure~\ref{fig:AvailableTimeSeries} and the heat-diffusion 1D partial differential equation defined in Appendix~\ref{s:Trafo}.}
	\label{fig:PDEmatlab}
\end{figure}

Figure~\ref{fig:PDEmatlab} shows the numerical solution of Eqs.~(\ref{eq:PDE_1D_Diffusion})--(\ref{eq:BoundaryConditions}), solved using finite element method (FEM) with Matlab's \texttt{pdepe} solver \cite{matlabpdepe}. The solution is experimentally validated against fiber optic sensor measurements installed at different transformer heights. Refer to \cite{Ramirez_25} for more details.

The main motivation for using PINNs over numerical methods is the computational effort and adaptability to different solutions. Numerical models require a mesh of parameters to model and evaluate the PDE. As for PINNs, there is no need to define the whole mesh.

Compared with classical data-driven methods, such as Neural Networks (NNs), PINNs enforce heat-diffusion equation throughout the domain, whereas classical NNs, only minimize the prediction error at measurement points. This improves the ability to generalize and operate with less data, compared with purely data-driven methods \cite{Ramirez_25}.

Furthermore, the Bayesian extension of PINNs enables the explicit quantification of epistemic uncertainty (model uncertainty due to limited data) and aleatoric uncertainty (measurement and process noise). This capability is particularly relevant for transformer lifetime assessment, where measurement deviations may lead to variations in thermal and ageing estimates. Providing uncertainty bounds allows operators to assess the reliability of predictions and supports risk-aware maintenance and operational decisions.

\subsection{Setup}

The implemented B-PINN configuration uses two layers with 50 neurons each, and Laplace prior distribution. An extensive hyperparameter study and justification of the selected parameters can be found in \cite{BPINN_PHM25}. The influence of the number of boundary condition points, initial points, and residual points is further evaluated in Subsection~\ref{ss:Ablation} to determine their influence on the accuracy and uncertainty of the solution. For the homoscedastic B-PINN configuration, input data and residual data variance follow $\sigma_i$=$\mathcal{N}(0,0.01)$ and $\sigma_r$=$\mathcal{N}(0,0.01)$.

All experiments use VI to approximate the posterior distribution. Each simulation is repeated five times for numerical stability, and mean and standard deviation values are reported. All models are implemented from scratch using TensorFlow Probability. The optimization algorithm is Adam, with a learning rate of 0.01 and a batch size of 16. All B-PINN models are trained for 15000 epochs, with early stopping applied using a patience of 200 epochs. The weights of the individual loss terms have been manually weighted, with final values of $\lambda_0$=1, $\lambda_b$=1, and $\lambda_r$=10\textsuperscript{-4}.

The vanilla PINN model is implemented using the Adam optimization algorithm for the first 20000 iterations, followed by the L-BFGS optimizer for the last 10000 iterations. The terms of individual loss terms have been manually weighted with final values of $\lambda_0$=1, $\lambda_b$=1, and $\lambda_r$=10\textsuperscript{-6}. Finally, the d-PINN model is designed taking the vanilla PINN architecture as the reference model, including additional dropout layers. The dropout rate is tested for a range of values $\rho$=[0.05, 0.1, 0.15, 0.2, 0.25], with $K$=200 Monte Carlo iterations. Further details of the benchmarking models are provided in Appendix~\ref{s:benchmarking_metrics}. 

Table \ref{tab:Hyperparameters} synthesizes the main parameters of the benchmarked models. The main differences arise from the employed underlying optimization algorithm, along with the selection of the dropout-rate for the d-PINN model and the selection of a prior distribution for B-PINNs. In all three configurations, no automatic loss weight adaptation strategy has been adopted, and they have been set manually (see Subsection~\ref{ss:BPINN_Training} for a extended discussion).

\begin{table}[!ht]
\centering
\caption{Configurations and parameters of the tested models.}
\label{tab:Hyperparameters}
\begin{tabular}{lccc}\hline
\textbf{Parameter} & \textbf{B-PINN} & \textbf{d-PINN} & \textbf{vanilla-PINN} \\ \hline
\# Layers & 2 & 2  & 2 \\ \hline
\# Neurons & 50 &  50 &  50 \\ \hline
Prior & Laplace & -  & -  \\ \hline
$N_b$ & 5760 & 5760  & 5760  \\ \hline
$N_r$ & 10000 & 10000  & 10000  \\ \hline
$N_i$ & 200 & 200  &  200 \\ \hline
Loss & VI &  NLL &  MSE \\ \hline
Optimization & Gradient Descent (Adam) &  Gradient Descent (Adam) &  Gradient Descent (Adam) \\ \hline
\end{tabular}
\end{table}

\subsection{Results}
\label{ss:Results}

\subsubsection*{Comparative Results}
\label{ss:Comparative_Results}

Five main configurations are evaluated in this study: (a) B-PINN with total uncertainty quantification (epistemic and heteroscedastic aleatoric uncertainty), (b) d-PINN with total uncertainty quantification (epistemic and heteroscedastic aleatoric uncertainty), (c)  B-PINN with epistemic uncertainty and homoscedastic aleatoric noise, (d) d-PINN with epistemic uncertainty and homoscedastic aleatoric noise, and (e) vanilla deterministic PINN without uncertainty quantification. 

Configurations (a) and (b) enable a direct comparison of total UQ capabilities between B-PINN and d-PINN. The impact of modelling heteroscedastic aleatoric uncertainty is subsequently analysed by comparing configurations (a) and (b) with their homoscedastic counterparts (c) and (d). The vanilla PINN serves as a deterministic baseline for completeness.

Table~\ref{tab:TotalResults} reports the probabilistic performance metrics (defined in Appendix \ref{s:benchmarking_metrics}) across different prediction instants for all considered configurations, including  continuously ranked probability score (CRPS) and negative log-likelihood, along with root-mean-square error (RMSE). Overall, the heteroscedastic B-PINN achieves the best performance across all metrics and time instants. In particular, it achieves a total RMSE of $1.09$, corresponding to a $57.3\%$ improvement over the vanilla PINN. When compared with the heteroscedastic d-PINN, the B-PINN yields a $42.6\%$ reduction in RMSE and a $40.2\%$ improvement in CRPS, highlighting its superior uncertainty calibration and predictive accuracy.

\begin{table}[!htb]
\small
\centering
\caption{Comparison of probabilistic performance results for different prediction instants across different B-PINN and d-PINN models with heteroscedastic and homoscedastic UQ, along with epistemic UQ. For completeness, vanilla PINN results are also reported. Values reported as mean$\pm$std, best results highlighted.}
\label{tab:TotalResults}
\setlength{\tabcolsep}{2.5pt}
\renewcommand{\arraystretch}{1.45}
\resizebox{\columnwidth}{!}{
\begin{tabular}{c|ccc|ccc|ccc|ccc|c|}
\hline
\multirow{3}{*}{\textbf{Instant [h]}} & \multicolumn{6}{c|}{\textbf{Heteroscedastic UQ}} 
 & \multicolumn{6}{c|}{\textbf{Homoscedastic UQ}} 
 & \multirow{2}{*}{\textbf{Vanilla PINN}} \\ \cline{2-13}
 & \multicolumn{3}{c|}{\textbf{B-PINN}} 
 & \multicolumn{3}{c|}{\textbf{d-PINN}} 
 & \multicolumn{3}{c|}{\textbf{B-PINN}} 
 & \multicolumn{3}{c|}{\textbf{d-PINN}}
 & \\
\cline{2-14}
& RMSE($\downarrow$) & CRPS($\downarrow$) & NLL($\downarrow$)
& RMSE($\downarrow$) & CRPS($\downarrow$) & NLL($\downarrow$)
& RMSE($\downarrow$) & CRPS($\downarrow$) & NLL($\downarrow$)
& RMSE($\downarrow$) & CRPS($\downarrow$) & NLL($\downarrow$)
& RMSE($\downarrow$)\\
\hline

0
& $0.03$\textsubscript{{\footnotesize $\pm$0.02}}& \cellcolor{selected}$0.03$\textsubscript{{\footnotesize $\pm$0.01}}& \cellcolor{selected}$-1.25$\textsubscript{{\footnotesize $\pm$0.45}}
& $0.14$\textsubscript{{\footnotesize $\pm$0.00}} & $0.23$\textsubscript{{\footnotesize $\pm$0.02}} & $0.84$\textsubscript{{\footnotesize $\pm$0.06}}
& \cellcolor{selected} $0.02$\textsubscript{{\footnotesize $\pm$0.01}} & $0.12$\textsubscript{{\footnotesize $\pm$0.08}} & $0.16$\textsubscript{{\footnotesize $\pm$0.69}}
& $0.20$\textsubscript{{\footnotesize $\pm$0.02}} & $0.38$\textsubscript{{\footnotesize $\pm$0.01}} & $1.38$\textsubscript{{\footnotesize $\pm$0.02}} 
& 1.47\\
\hline

3
& \cellcolor{selected}$0.31$\textsubscript{{\footnotesize $\pm$0.04}}& \cellcolor{selected}$0.19$\textsubscript{{\footnotesize $\pm$0.01}}& $0.41$\textsubscript{{\footnotesize $\pm$0.65}}
& $1.65$\textsubscript{{\footnotesize $\pm$0.25}} & $0.96$\textsubscript{{\footnotesize $\pm$0.12}} & $1.92$\textsubscript{{\footnotesize $\pm$0.01}}
& $0.33$\textsubscript{{\footnotesize $\pm$0.12}} & $0.20$\textsubscript{{\footnotesize $\pm$0.09}} &\cellcolor{selected} $0.40$\textsubscript{{\footnotesize $\pm$0.54}}
& $1.77$\textsubscript{{\footnotesize $\pm$0.14}} & $1.06$\textsubscript{{\footnotesize $\pm$0.10}} & $2.00$\textsubscript{{\footnotesize $\pm$0.10}} 
& 1.45 \\
\hline

6
& \cellcolor{selected}$0.32$\textsubscript{{\footnotesize $\pm$0.04}}& \cellcolor{selected}$0.21$\textsubscript{{\footnotesize $\pm$0.04}}& $1.61$\textsubscript{{\footnotesize $\pm$1.43}}
& $2.15$\textsubscript{{\footnotesize $\pm$0.31}} & $1.24$\textsubscript{{\footnotesize $\pm$0.18}} & $2.13$\textsubscript{{\footnotesize $\pm$0.05}}
& $0.30$\textsubscript{{\footnotesize $\pm$0.01}} & $0.20$\textsubscript{{\footnotesize $\pm$0.03}} &  \cellcolor{selected}$0.49$\textsubscript{{\footnotesize $\pm$0.33}}
& $1.86$\textsubscript{{\footnotesize $\pm$0.20}} & $1.13$\textsubscript{{\footnotesize $\pm$0.14}} & $2.07$\textsubscript{{\footnotesize $\pm$0.15}}
& 2.17\\
\hline

18
& \cellcolor{selected}$0.34$\textsubscript{{\footnotesize $\pm$0.01}}& \cellcolor{selected}$0.22$\textsubscript{{\footnotesize $\pm$0.01}}& \cellcolor{selected}$0.63$\textsubscript{{\footnotesize $\pm$0.12}}
& $1.30$\textsubscript{{\footnotesize $\pm$0.06}} & $0.88$\textsubscript{{\footnotesize $\pm$0.15}} & $1.94$\textsubscript{{\footnotesize $\pm$0.36}}
& $0.67$\textsubscript{{\footnotesize $\pm$0.39}} & $0.41$\textsubscript{{\footnotesize $\pm$0.23}} & $1.07$\textsubscript{{\footnotesize $\pm$0.57}}
& $1.16$\textsubscript{{\footnotesize $\pm$0.81}} & $0.61$\textsubscript{{\footnotesize $\pm$0.32}} & $1.59$\textsubscript{{\footnotesize $\pm$0.28}} 
& 3.29\\
\hline

25
& \cellcolor{selected}$0.42$\textsubscript{{\footnotesize $\pm$0.02}}& \cellcolor{selected}$0.26$\textsubscript{{\footnotesize $\pm$0.02}}& \cellcolor{selected}$0.93$\textsubscript{{\footnotesize $\pm$0.09}}
& $2.04$\textsubscript{{\footnotesize $\pm$0.47}} & $1.18$\textsubscript{{\footnotesize $\pm$0.31}} & $2.12$\textsubscript{{\footnotesize $\pm$0.37}}
& $0.57$\textsubscript{{\footnotesize $\pm$0.05}} & $0.35$\textsubscript{{\footnotesize $\pm$0.05}} &  $1.39$\textsubscript{{\footnotesize $\pm$0.71}}
& $1.81$\textsubscript{{\footnotesize $\pm$0.11}} & $1.07$\textsubscript{{\footnotesize $\pm$0.07}} & $2.01$\textsubscript{{\footnotesize $\pm$0.07}} 
& 1.78\\
\hline

50
& \cellcolor{selected}$0.45$\textsubscript{{\footnotesize $\pm$0.13}}& \cellcolor{selected}$0.28$\textsubscript{{\footnotesize $\pm$0.08}}& \cellcolor{selected}$0.72$\textsubscript{{\footnotesize $\pm$0.28}}
& $3.35$\textsubscript{{\footnotesize $\pm$0.58}} & $2.06$\textsubscript{{\footnotesize $\pm$0.51}} & $2.76$\textsubscript{{\footnotesize $\pm$0.45}}
& $1.07$\textsubscript{{\footnotesize $\pm$0.83}} & $0.71$\textsubscript{{\footnotesize $\pm$0.58}} & $2.06$\textsubscript{{\footnotesize $\pm$1.33}}
& $3.47$\textsubscript{{\footnotesize $\pm$0.79}} & $0.89$\textsubscript{{\footnotesize $\pm$0.43}} & $1.86$\textsubscript{{\footnotesize $\pm$0.36}}
& 2.27 \\
\bottomrule

\textbf{Total}
& \cellcolor{selected}$1.09$\textsubscript{{\footnotesize $\pm$0.21}} & \cellcolor{selected}$0.58$\textsubscript{{\footnotesize $\pm$0.09}} & \cellcolor{selected}$1.75$\textsubscript{{\footnotesize $\pm$0.19}}
& $1.90$\textsubscript{{\footnotesize $\pm$0.57}} & $0.97$\textsubscript{{\footnotesize $\pm$0.36}} & $1.80$\textsubscript{{\footnotesize $\pm$0.45}}
& $1.12$\textsubscript{{\footnotesize $\pm$0.47}} & $0.66$\textsubscript{{\footnotesize $\pm$0.23}} & $1.98$\textsubscript{{\footnotesize $\pm$0.73}}
& $1.96$\textsubscript{{\footnotesize $\pm$0.13}} & $1.10$\textsubscript{{\footnotesize $\pm$0.06}} & $2.10$\textsubscript{{\footnotesize $\pm$0.07}}
& 2.55\\
\bottomrule

\end{tabular}}
\end{table}

The heteroscedastic B-PINN formulation consistently outperforms its homoscedastic counterpart across different prediction instants. While total RMSE values remain comparable between both variants, remarkable improvements are observed in CRPS, indicating better calibrated predictive distributions when heteroscedastic uncertainty is modelled. Although the homoscedastic B-PINN achieves slightly lower NLL values at specific instants (e.g., $t=6$\,h), the heteroscedastic B-PINN yields an overall NLL improvement of $11.6\%$. At early instants, the heteroscedastic B-PINN achieves near-zero errors, while maintaining stable performance as the prediction instant increases.

In contrast, d-PINN exhibit significantly degraded performance, particularly at longer horizons. For instance, at $t=50$\,h, the heteroscedastic d-PINN reaches an RMSE of $3.35$, representing a $645\%$ increase relative to the heteroscedastic B-PINN. This degradation reflects reduced predictive accuracy and over-dispersed uncertainty estimates, underscoring the limitations of dropout-based uncertainty approximation in the analysed setting.

\begin{figure}[!htb]
	\centering
	\includegraphics[width=0.43\columnwidth]{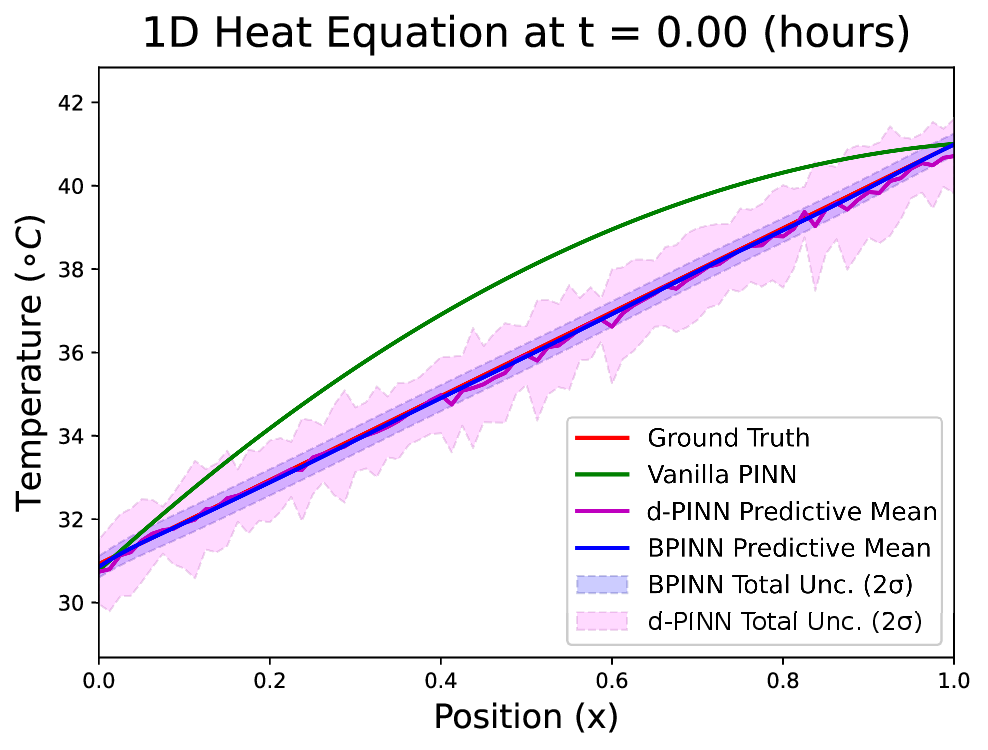}
	\includegraphics[width=0.43\columnwidth]{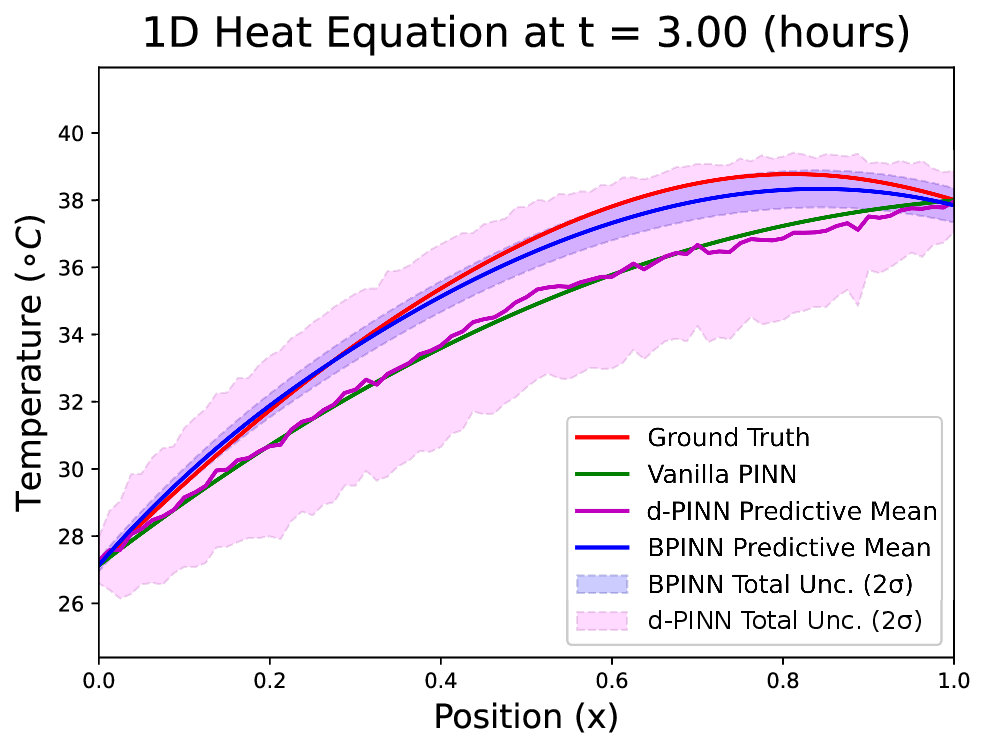}\\[0.5em]
	\includegraphics[width=0.43\columnwidth]{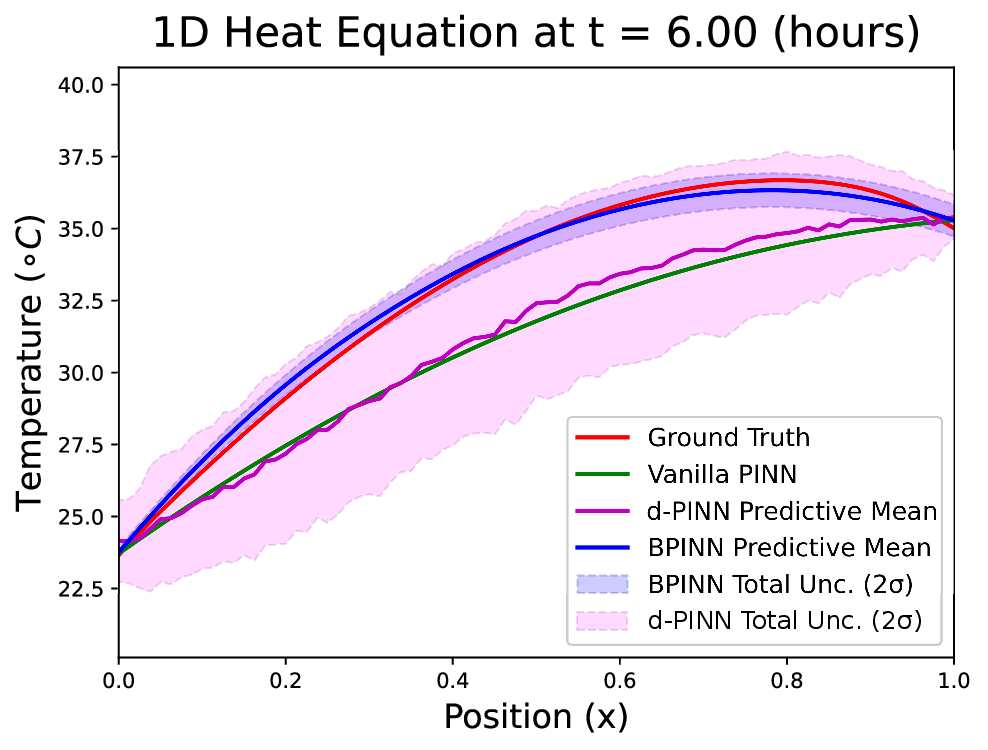}
	\includegraphics[width=0.43\columnwidth]{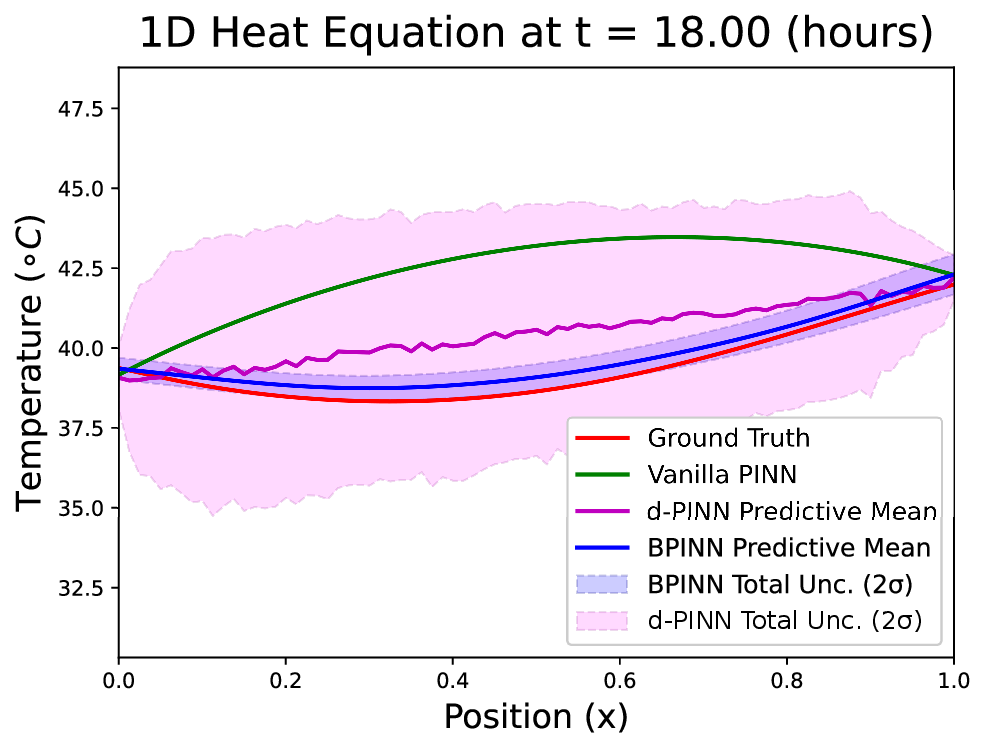}\\[0.5em]
	\includegraphics[width=0.43\columnwidth]{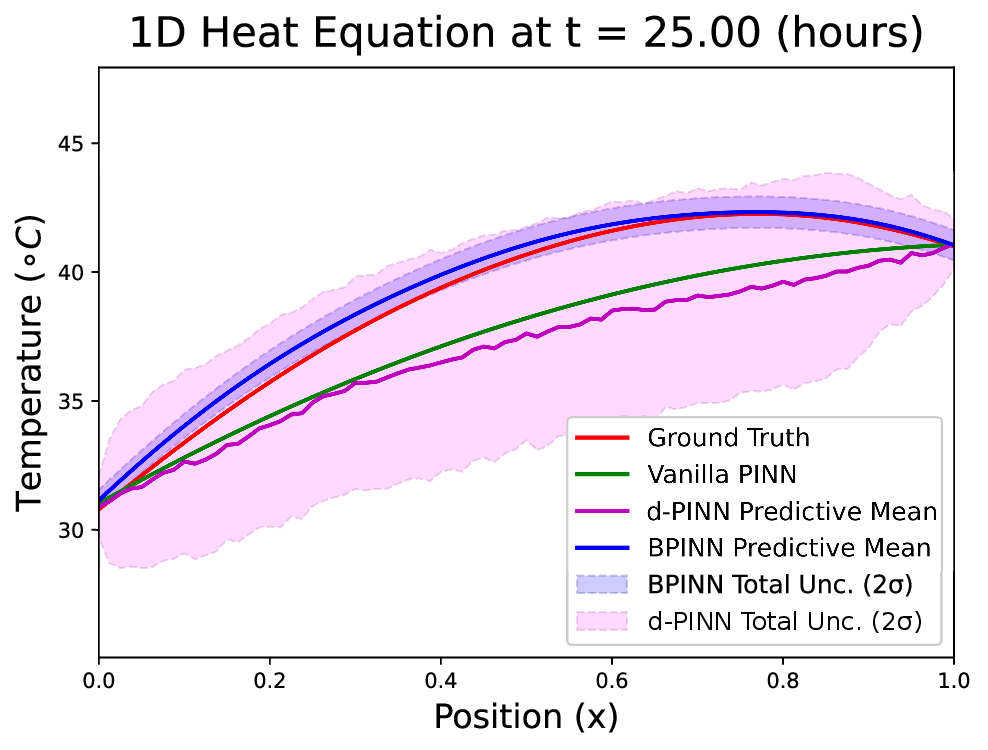}
	\includegraphics[width=0.43\columnwidth]{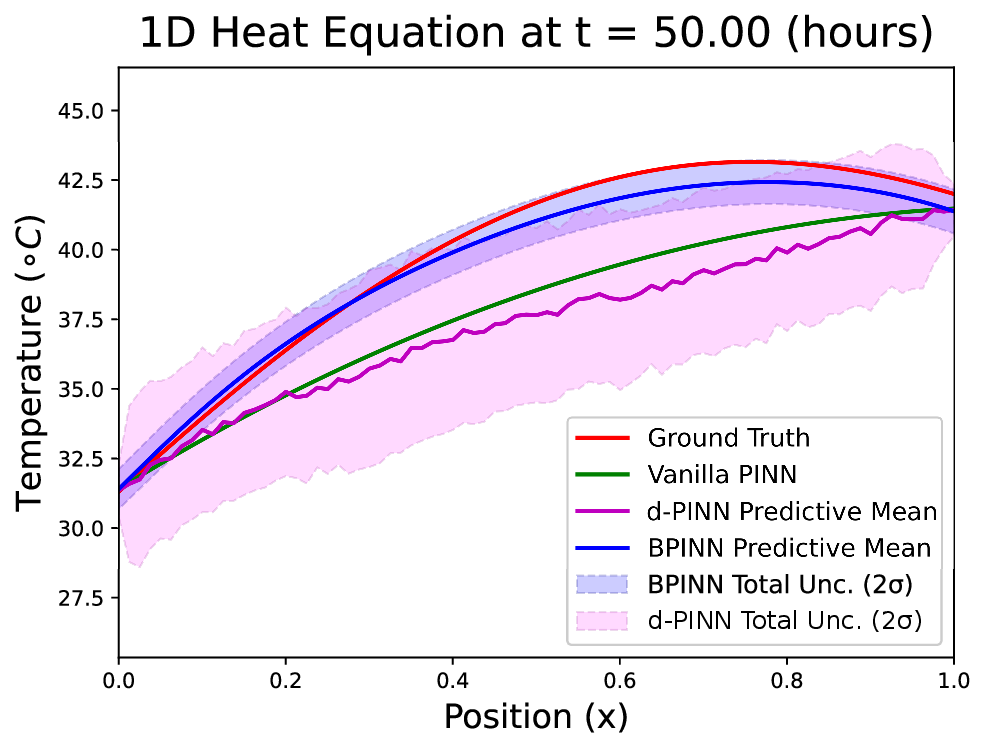}
	\caption{Spatiotemporal transformer insulation oil temperature estimates at different time instants for heteroscedastic B-PINN, heteroscedastic d-PINN, and vanilla PINN models, along with the corresponding measured ground truth values.}
	\label{fig:TemperatureEstimation_Instants}
\end{figure}

Figure~\ref{fig:TemperatureEstimation_Instants} illustrates the spatiotemporal transformer insulation oil temperature predictions at selected time instants for the heteroscedastic B-PINN, heteroscedastic d-PINN, and vanilla PINN models, together with the measured ground truth.

Among the heteroscedastic models, the total predictive uncertainty produced by the d-PINN is consistently wider than that of the B-PINN across all considered time instants and spatial locations. This reflects the approximate nature of Monte Carlo dropout, which estimates epistemic uncertainty through stochastic neuron deactivation \cite{BPINN_PHM25}.  As the prediction instant increases and extrapolation becomes more pronounced, the variability between stochastic forward passes increases, resulting in wider uncertainty intervals. In contrast, the B-PINN learns structured posterior weight distributions constrained by physical laws and prior information, leading to more stable extrapolation and better-calibrated uncertainty estimates.

In contrast, the B-PINN yields tighter and more structured uncertainty bands while maintaining closer agreement between the predictive mean and the ground truth temperature profiles. The predictive mean of the d-PINN closely follows that of the vanilla PINN, with small but systematic deviations from the measured values. This indicates that dropout-based uncertainty quantification primarily affects the dispersion of predictions rather than substantially correcting the bias of the underlying deterministic solution. Overall, these results highlight the advantages of fully Bayesian inference in B-PINNs for achieving a more balanced trade-off between predictive accuracy and uncertainty sharpness.

\subsubsection*{Uncertainty Assessment}
\label{ss:Uncertainty_Assessment}

Focusing on the homoscedastic and heteroscedastic B-PINN configurations that incorporate epistemic uncertainty, Table~\ref{tab:CalibSharpResults} displays uncertainty-specific metrics that quantify both calibration and sharpness. Calibration reflects the agreement between empirical coverage and nominal confidence levels, measured through the miscalibration area, while sharpness characterizes the concentration of the predictive distributions independently of calibration.

\begin{table}[!ht]
\centering
\caption{Comparison of calibration and sharpness metrics for homoscedastic and heteroscedastic B-PINN models evaluated across the operational days shown in Figure~\ref{fig:AvailableTimeSeries}.}
\label{tab:CalibSharpResults}
\begin{tabular}{lcc}
\hline
\textbf{Metric} & \textbf{B-PINN Heteroscedastic} & \textbf{B-PINN Homoscedastic} \\
\hline
Miscalibration area & $0.1956 \pm 0.1616$ &  $0.2135 \pm 0.0591$\\
Sharpness           & $0.5258 \pm 0.3369$  & $0.5603 \pm 0.0892$ \\
\hline
\end{tabular}
\end{table}

From Table~\ref{tab:CalibSharpResults}, it can be observed that the heteroscedastic B-PINN achieves a slightly lower miscalibration area compared with the homoscedastic configuration, indicating marginally improved calibration of the predictive uncertainty. This suggests that allowing the aleatoric variance to depend on the input data enables the B-PINN to better adapt uncertainty levels across the spatiotemporal domain. 

In terms of sharpness, the heteroscedastic B-PINN produces slightly sharper predictive distributions on average, but with increased variability. This reflects localized increases in uncertainty  within regions characterized by higher model uncertainty or data uncertainty. Overall, these results highlight the inherent trade-off between calibration and sharpness and demonstrate that heteroscedastic modeling provides a more flexible and expressive uncertainty representation when combined with epistemic uncertainty in Bayesian PINNs.

\begin{figure*}[!ht]
    \centering
    \begin{subfigure}[b]{0.49\textwidth}
        \centering
        \includegraphics[width=\textwidth]{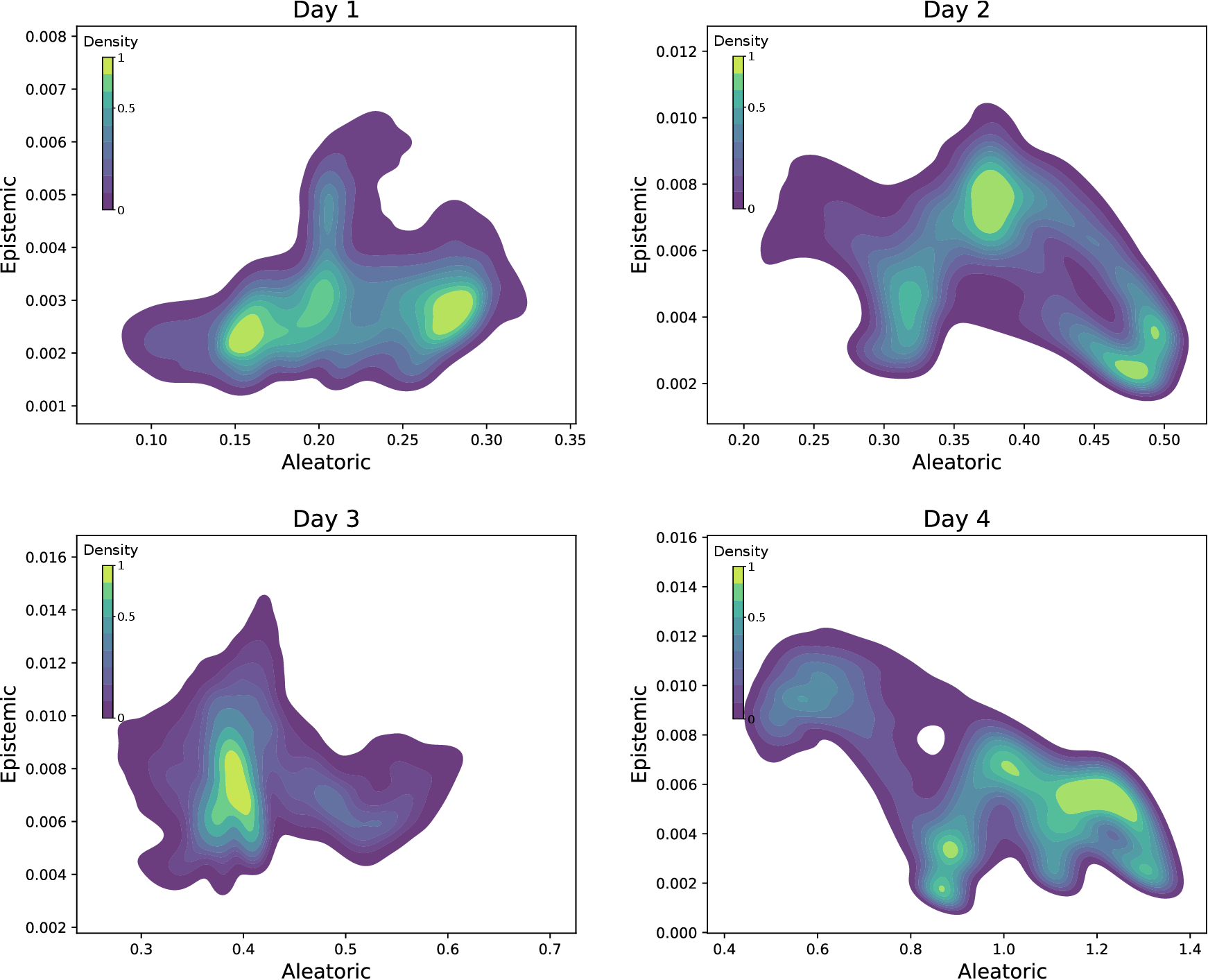}
        \caption{B-PINN model results.}
    \end{subfigure}
    \begin{subfigure}[b]{0.49\textwidth}
        \centering
        \includegraphics[width=\textwidth]{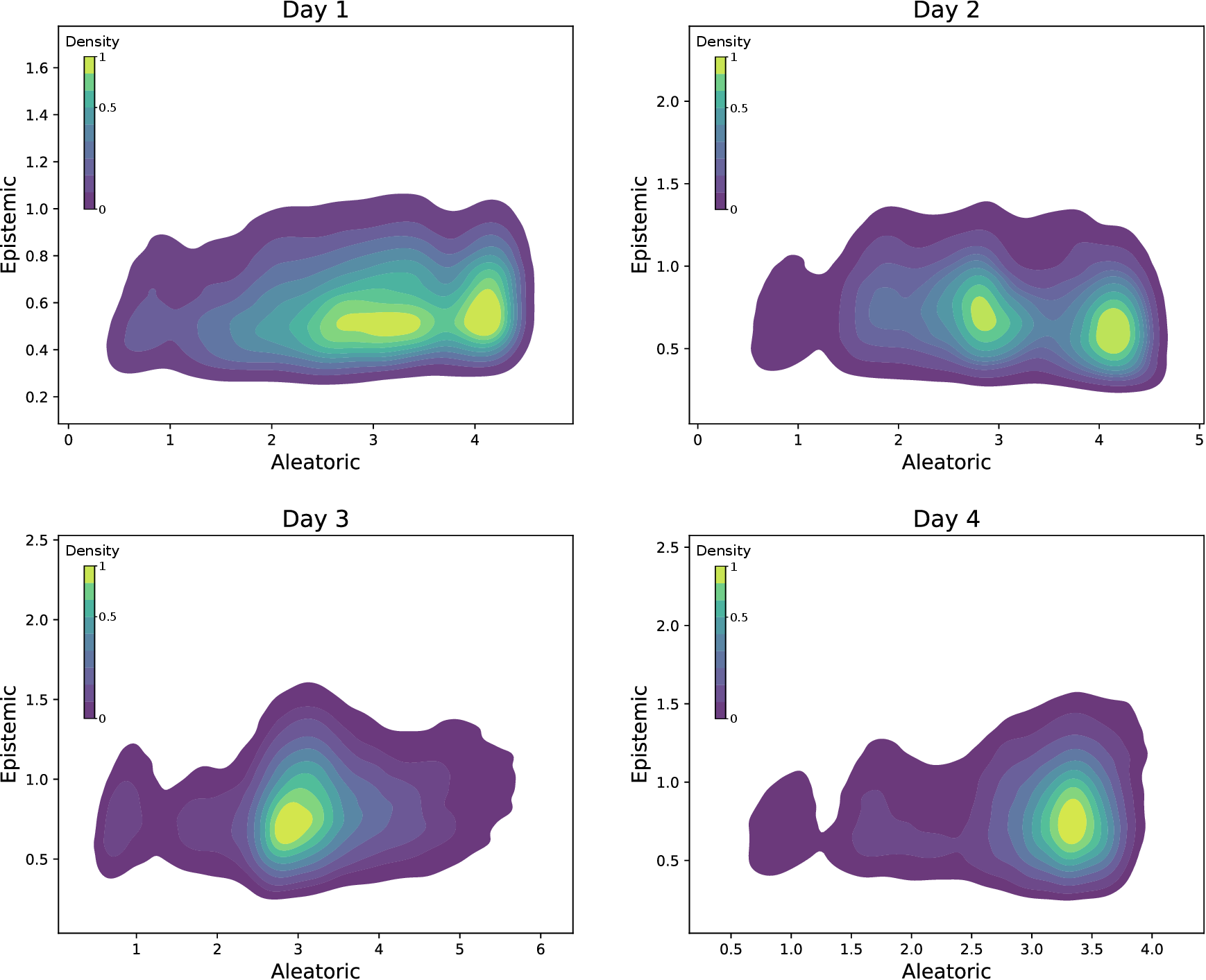}
        \caption{d-PINN model results.}
    \end{subfigure}
    \caption{Daily joint probability density of aleatoric (heteroscedastic) and epistemic uncertainty estimated by the (a) B-PINN model and (b) d-PINN model. The color intensity indicates the relative density of uncertainty realizations, with brighter regions corresponding to higher probability mass.}
    \label{fig:ale_epis_density}
\end{figure*}

Taking the heteroscedastic B-PINN model as a reference, the relative contribution of epistemic and aleatoric uncertainty is analyzed through their daily joint probability densities. Figure~\ref{fig:ale_epis_density}(a) presents the joint distributions for four representative days.

For the B-PINN, the highest-density regions are consistently concentrated around well-defined modes in the joint aleatoric–epistemic space, indicating stable and dominant uncertainty regimes. Across all considered days, epistemic uncertainty remains confined to a relatively narrow interval of approximately $\langle 0.001, 0.016 \rangle$, whereas aleatoric uncertainty spans a broader range of approximately $\langle 0.10, 1.4 \rangle$. This behavior reflects the input-dependent nature of the heteroscedastic noise formulation, while maintaining controlled epistemic variability.

Day-to-day variations are nonetheless observable. The dispersion of epistemic uncertainty is most pronounced on days 3 and 4, suggesting increased model uncertainty potentially associated with reduced data support or higher physical model mismatch. Aleatoric uncertainty exhibits its largest variability on day 4, indicating stronger input-dependent noise effects under that specific operating condition. Importantly, the compact concentration of probability mass across all days implies that most uncertainty realizations remain confined to narrow aleatoric–epistemic regimes, with occasional excursions into higher-uncertainty states. This indicates stable disentanglement between data-driven and model-driven uncertainty sources.

Figure~\ref{fig:ale_epis_density}(b) shows the corresponding joint densities obtained using the heteroscedastic d-PINN model. In contrast to the B-PINN results, both uncertainty components exhibit substantially larger magnitudes and broader dispersion. Epistemic uncertainty spans approximately $\langle 0.2, 1.8 \rangle$, while aleatoric uncertainty varies over an extended interval of roughly $\langle 0.5, 5.5 \rangle$. The joint densities appear more diffuse and horizontally elongated, particularly on days 3 and 4, indicating increased epistemic variability. Although input-dependent aleatoric behavior is still present, the broader spread suggests a less stable separation between epistemic and aleatoric contributions under dropout-based approximation.

A direct comparison of Figures~\ref{fig:ale_epis_density}(a) and (b) highlights the key distinction between the two approaches. The B-PINN produces lower-magnitude and more concentrated uncertainty regimes, with well-structured joint distributions and limited dispersion tails. In contrast, the d-PINN yields systematically larger and more distributed uncertainty realizations, particularly in the epistemic component.

\begin{figure}[!htb]
    \centering
     \begin{subfigure}[b]{0.49\textwidth}
        \includegraphics[width=\textwidth]{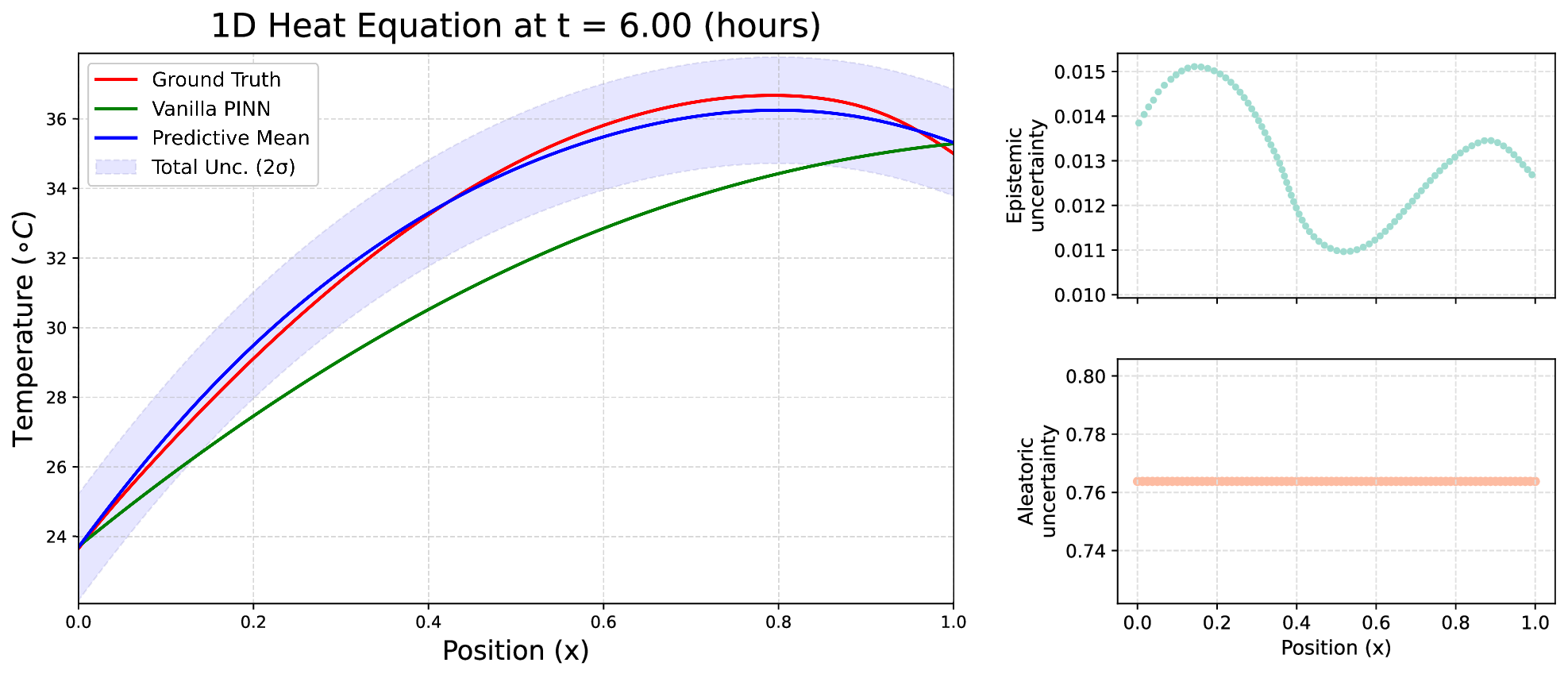}
        \caption{}
        \label{subfig:BPINN_homosc_6}
    \end{subfigure}
    \begin{subfigure}[b]{0.49\textwidth}
        \includegraphics[width=\textwidth]{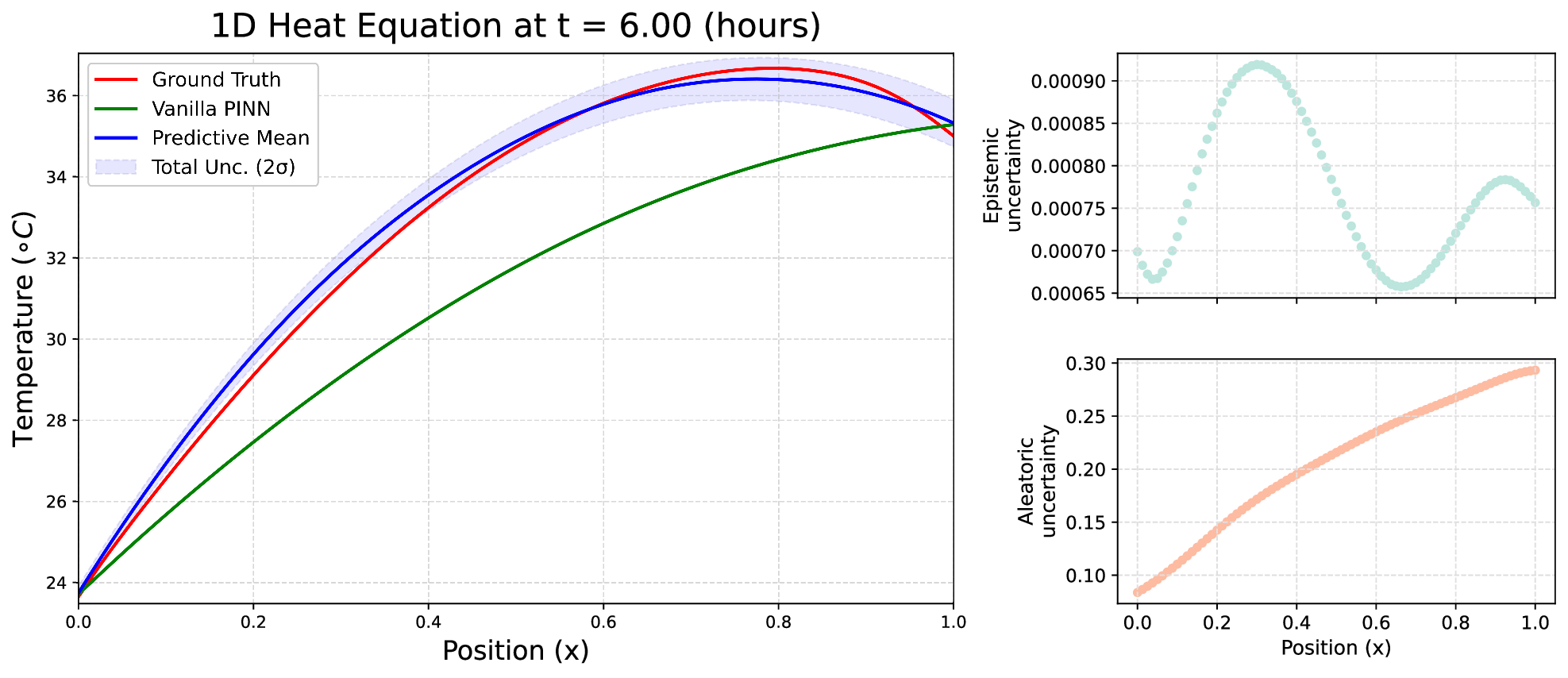}
        \caption{}
        \label{subfig:BPINN_heterosc_6}
    \end{subfigure}
    \begin{subfigure}[b]{0.49\textwidth}
        \includegraphics[width=\textwidth]{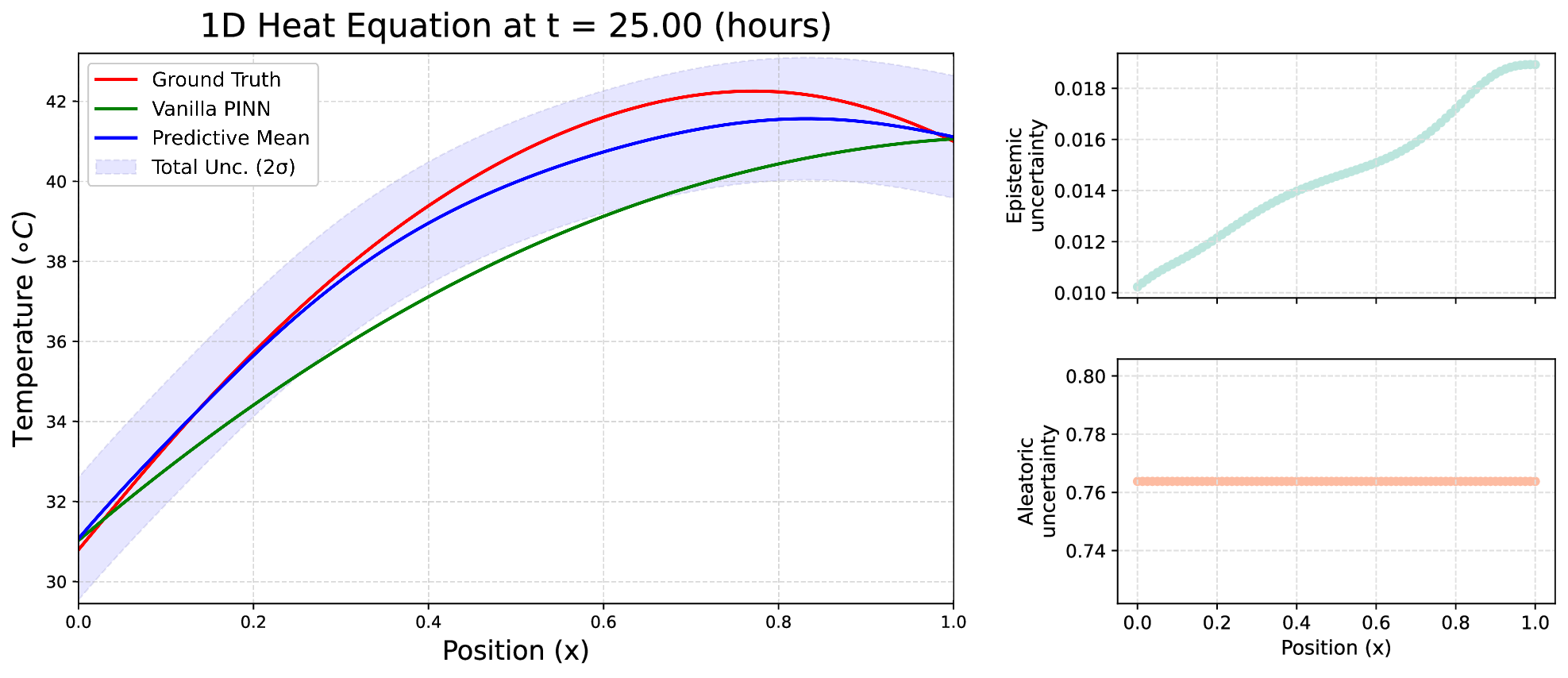}
        \caption{}
        \label{subfig:BPINN_homosc_25}
    \end{subfigure}
    \begin{subfigure}[b]{0.49\textwidth}
        \includegraphics[width=\textwidth]{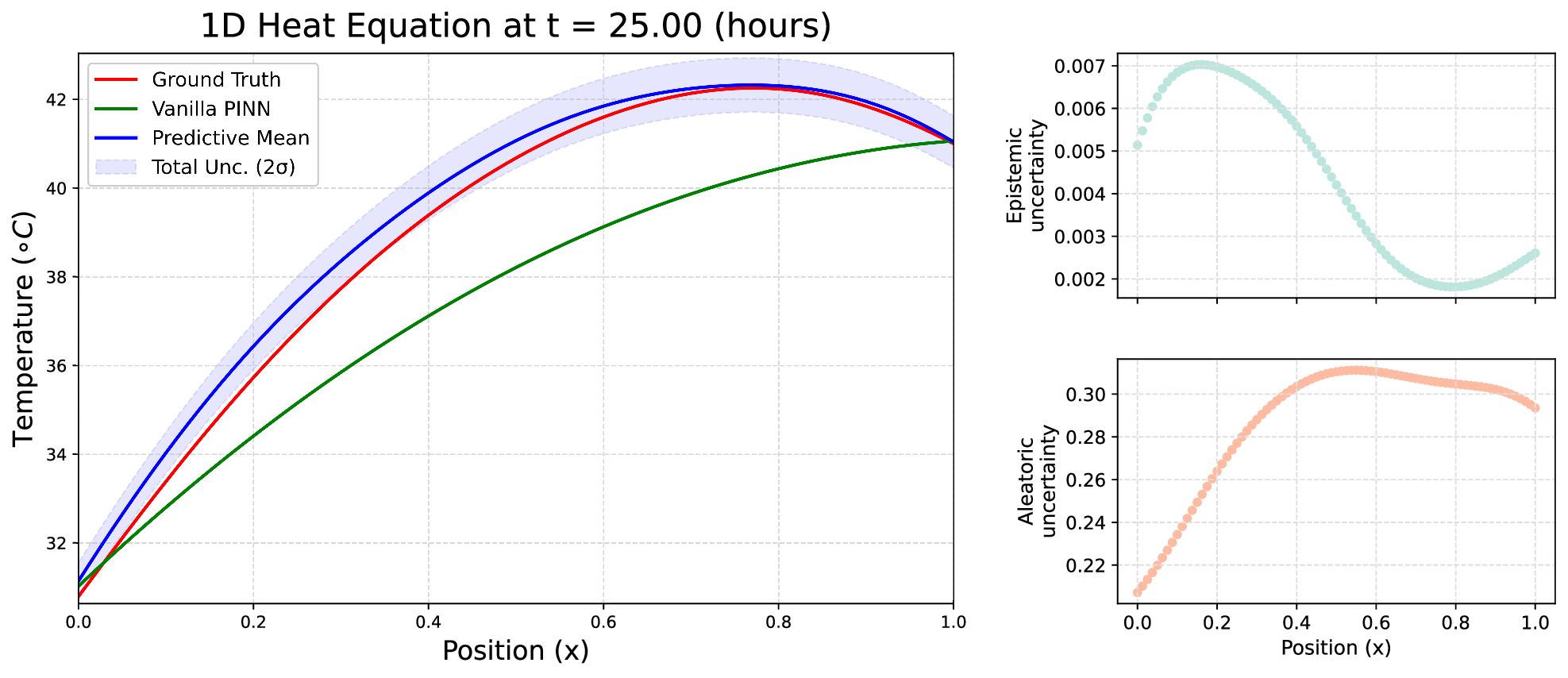}
        \caption{}
        \label{subfig:BPINN_heterosc_25}
    \end{subfigure}
    \caption{Spatiotemporal temperature estimates with the corresponding decomposition into epistemic and aleatoric uncertainty. (a) B-PINN homoscedastic at $t=6$ hours; (b) B-PINN heteroscedastic at $t=6$ hours; (c) B-PINN homoscedastic at $t=25$ hours; (b) B-PINN heteroscedastic at $t=25$ hours.}
    \label{fig:TemperatureEstimation_UQ}
\end{figure}

In order to further compare the relative contributions of epistemic and heteroscedastic aleatoric sources of uncertainty, the influence of each uncertainty component is quantified separately. Figure~\ref{fig:TemperatureEstimation_UQ} illustrates the measured ground truth, vanilla PINN output, and the proposed B-PINN mean and standard deviation predictions, along with the aleatoric and epistemic uncertainty estimates calculated by the B-PINN for the instants $t=6$ and $t=25$.

It can be observed from Figure~\ref{fig:TemperatureEstimation_UQ} that, in the homoscedastic B-PINN configuration, the aleatoric uncertainty remains spatially constant, reflecting the assumption of input-independent observation noise. As a result, the total predictive uncertainty exhibits a nearly uniform contribution from the aleatoric component across the spatial domain.

In contrast, the heteroscedastic B-PINN configuration learns an input-dependent aleatoric uncertainty, which adapts to variations in temperature. This input data leads to spatially varying uncertainty bands that more closely follow the underlying solution structure. Consequently, the heteroscedastic model yields uncertainty bounds that are generally tighter in well-informed regions while allowing localized uncertainty increase where the model or data uncertainty is higher. This comparison illustrates the increased expressiveness and reduced over-conservatism of heteroscedastic uncertainty modeling within the B-PINN framework.

Additionally, comparing the prediction error of the B-PINN model with the epistemic uncertainty shown in Figure~\ref{fig:TemperatureEstimation_UQ}(b) and Figure~\ref{fig:TemperatureEstimation_UQ}(d), it can be observed that epistemic uncertainty of the heteroscedastic configuration adequately tracks prediction errors of the B-PINN model. For the instant $t=6$ hours, the prediction error increases over two periods ($t=[0.1,0.45]$ and $t=[0.65,0.9]$) and is zero near $t=0.6$. For the instant $t=25$ hours, the prediction error decreases progressively until the instant  $t=0.75$. This is directly reflected in the epistemic uncertainty. However, note that due to the difference in the magnitued of epistemic and aleatoric uncertainties, this is not directly reflected in the total uncertainty.

\subsubsection*{Ageing Prediction}
\label{ss:Ageing}

\begin{figure}[!ht]
    \centering
    \begin{subfigure}[b]{0.49\textwidth}
        \centering
        \includegraphics[width=\textwidth]{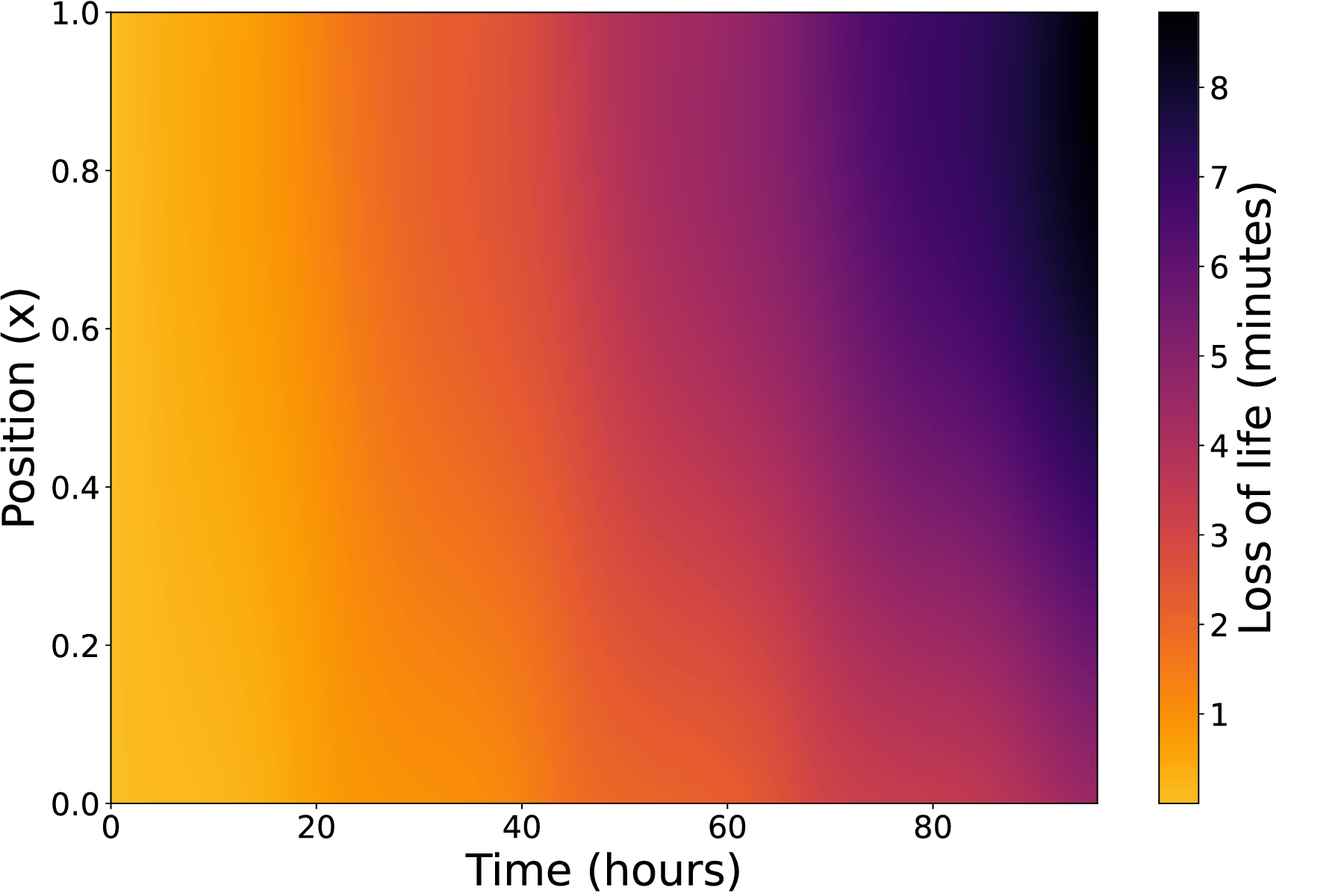}
        \caption{}
        \label{subfig:BPINN_mean}
    \end{subfigure}
    \begin{subfigure}[b]{0.49\textwidth}
        \centering
        \includegraphics[width=\textwidth]{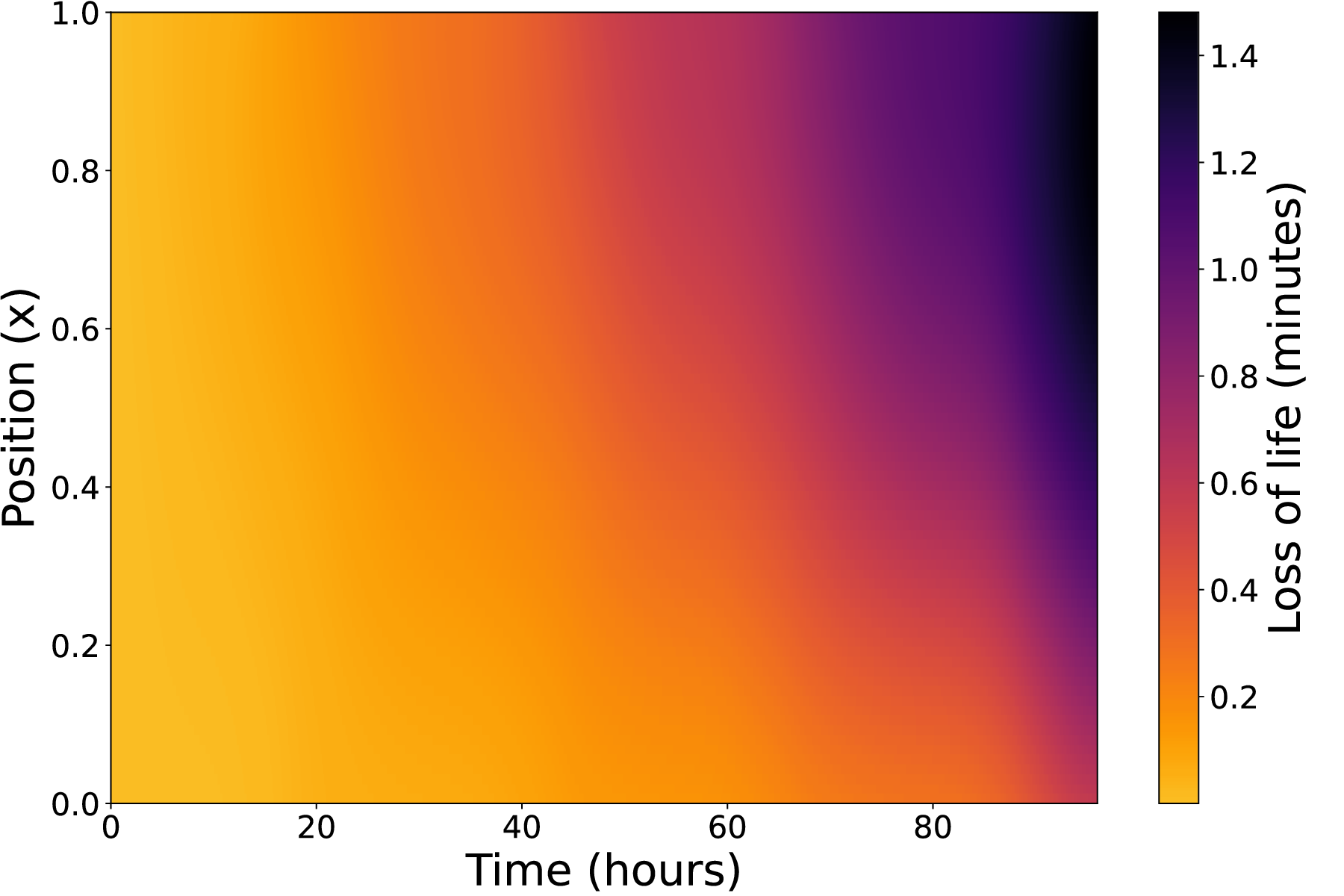 }
        \caption{}
        \label{subfig:BPINN_std}
    \end{subfigure}
    \vspace{0.5cm}
    \begin{subfigure}[b]{0.49\textwidth}
        \centering
        \includegraphics[width=\textwidth]{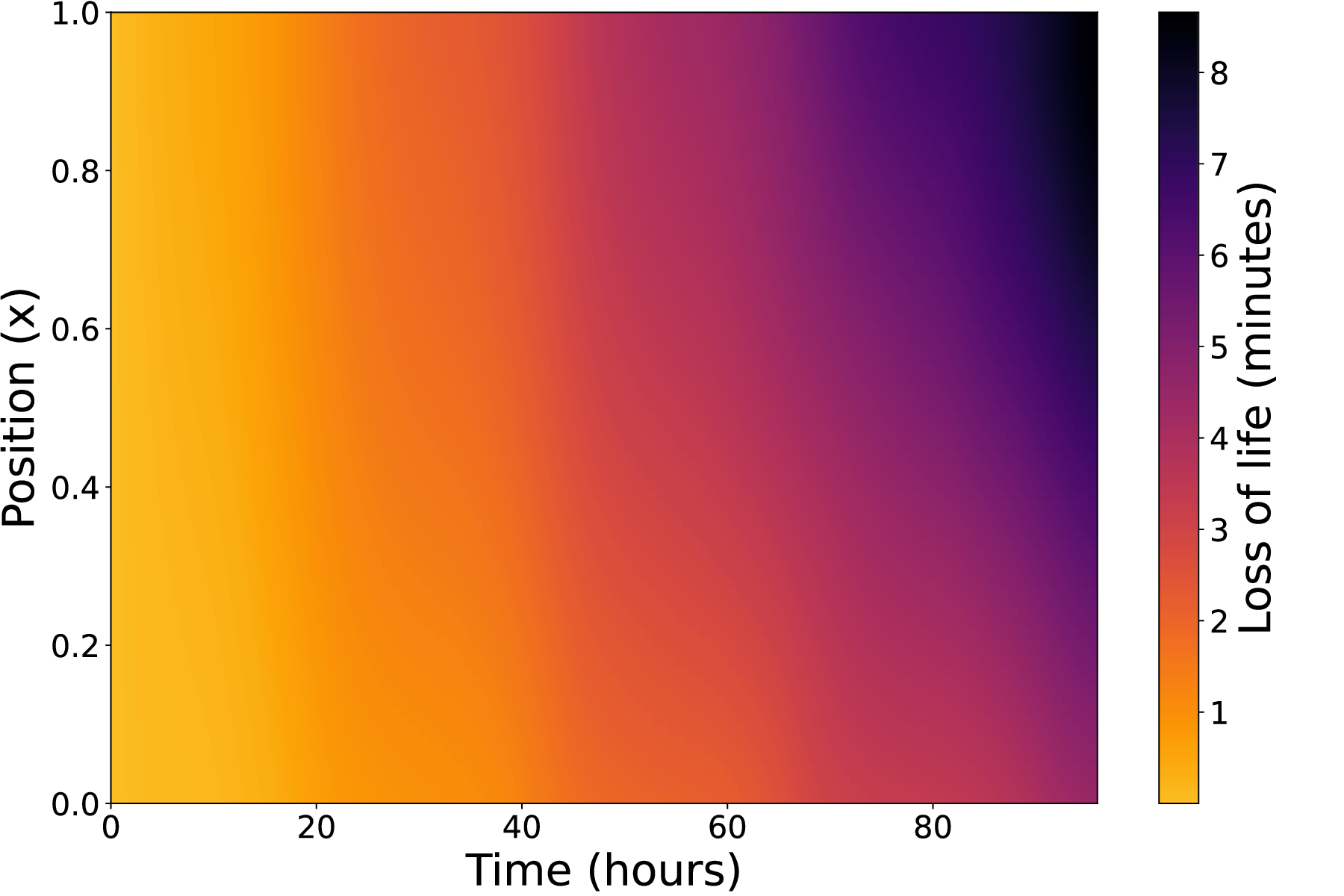}
        \caption{}
        \label{subfig:dPINN_mean}
    \end{subfigure}
    \begin{subfigure}[b]{0.49\textwidth}
        \centering
        \includegraphics[width=\textwidth]{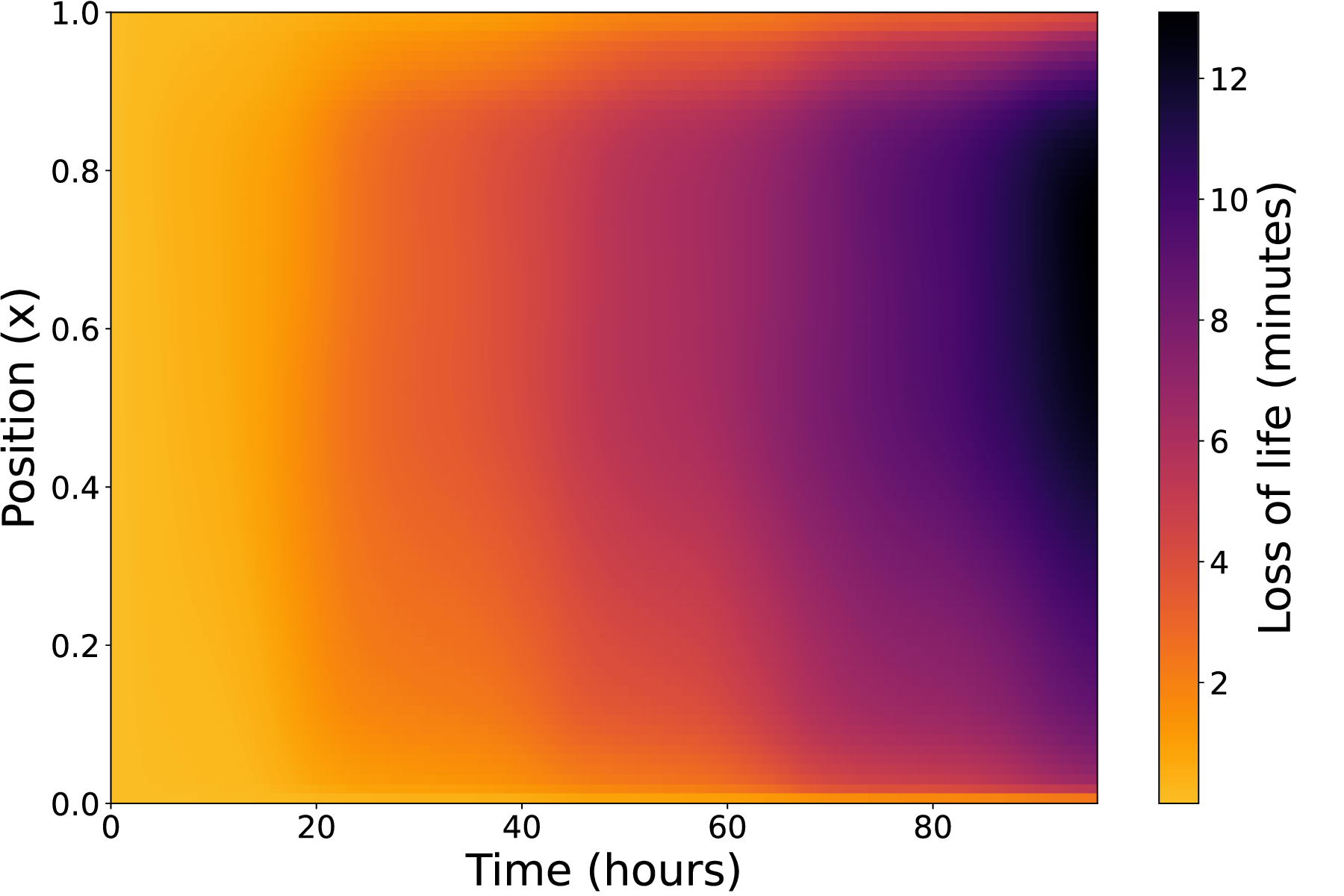}
        \caption{}
        \label{subfig:dPINN_std}
    \end{subfigure}
    \caption{Probabilistic spatiotemporal ageing predictions for B-PINN and d-PINN models with total UQ: (a) mean estimate of B-PINN, (b) standard deviation of B-PINN, (c) mean estimate of d-PINN, and (d) standard deviation of d-PINN.}
    \label{fig:heatmaps}
\end{figure}

Building on the spatiotemporal probabilistic temperature predictions obtained with the B-PINN and d-PINN models under total UQ, Figure~\ref{fig:heatmaps} shows the resulting  spatiotemporal insulation ageing estimates $p(\hat{V}(x,t))$. For each model, the predictive mean and the associated standard deviation are shown (see Appendix~\ref{ss:AgeingEstimate} for empirical ageing model details). 

Figures~\ref{fig:heatmaps}(a) and (b) illustrate the mean and standard deviation of the ageing estimates obtained with the B-PINN, while Figures~\ref{fig:heatmaps}(c) and (d) show the corresponding results for the d-PINN. The predictive mean ageing estimates are broadly consistent across both models, with maximum accumulated ageing values of 8.83 minutes for the B-PINN and 8.65 minutes for the d-PINN. This indicates that, at the level of point estimates, both models capture similar expected ageing dynamics.

However, substantial differences emerge in the associated predictive uncertainty. The B-PINN exhibits a maximum ageing standard deviation of 1.48 minutes, whereas the d-PINN achieves a larger maximum standard deviation of 13.09 minutes. This discrepancy reflects the significantly higher uncertainty propagated by the dropout-based approximation, particularly when integrated over space and time. In this regime, the standard deviation of the d-PINN ageing estimate exceeds the corresponding mean value, indicating a loss of informative predictive content and limited practical interpretability.

The differences between Figure~\ref{fig:heatmaps}(c) and (d) highlight the critical impact of uncertainty modeling choices in prognostics applications. The wider uncertainty bands produced by the d-PINN indicate a tendency toward overly conservative ageing estimates. When employed in maintenance and asset management decision-making, such uncertainty may lead to premature asset replacement, even in scenarios where the expected ageing remains moderate.

\begin{figure}[!ht]
    \centering
    \begin{subfigure}[b]{0.49\textwidth}
        \centering
        \includegraphics[width=\textwidth]{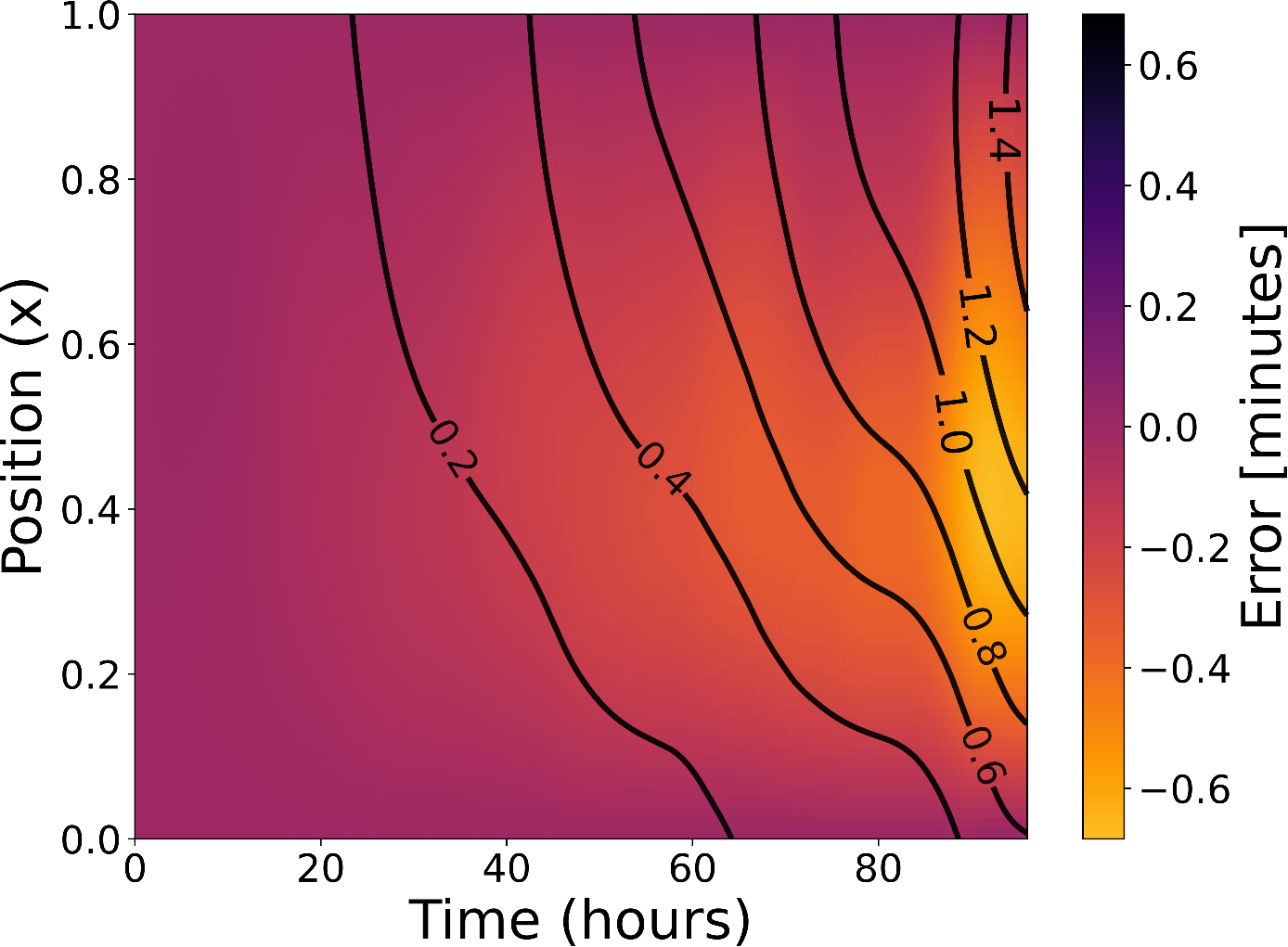}
        \caption{}
        \label{subfig:BPINN_error}
    \end{subfigure}
    \hfill
    \begin{subfigure}[b]{0.49\textwidth}
        \centering
        \includegraphics[width=\textwidth]{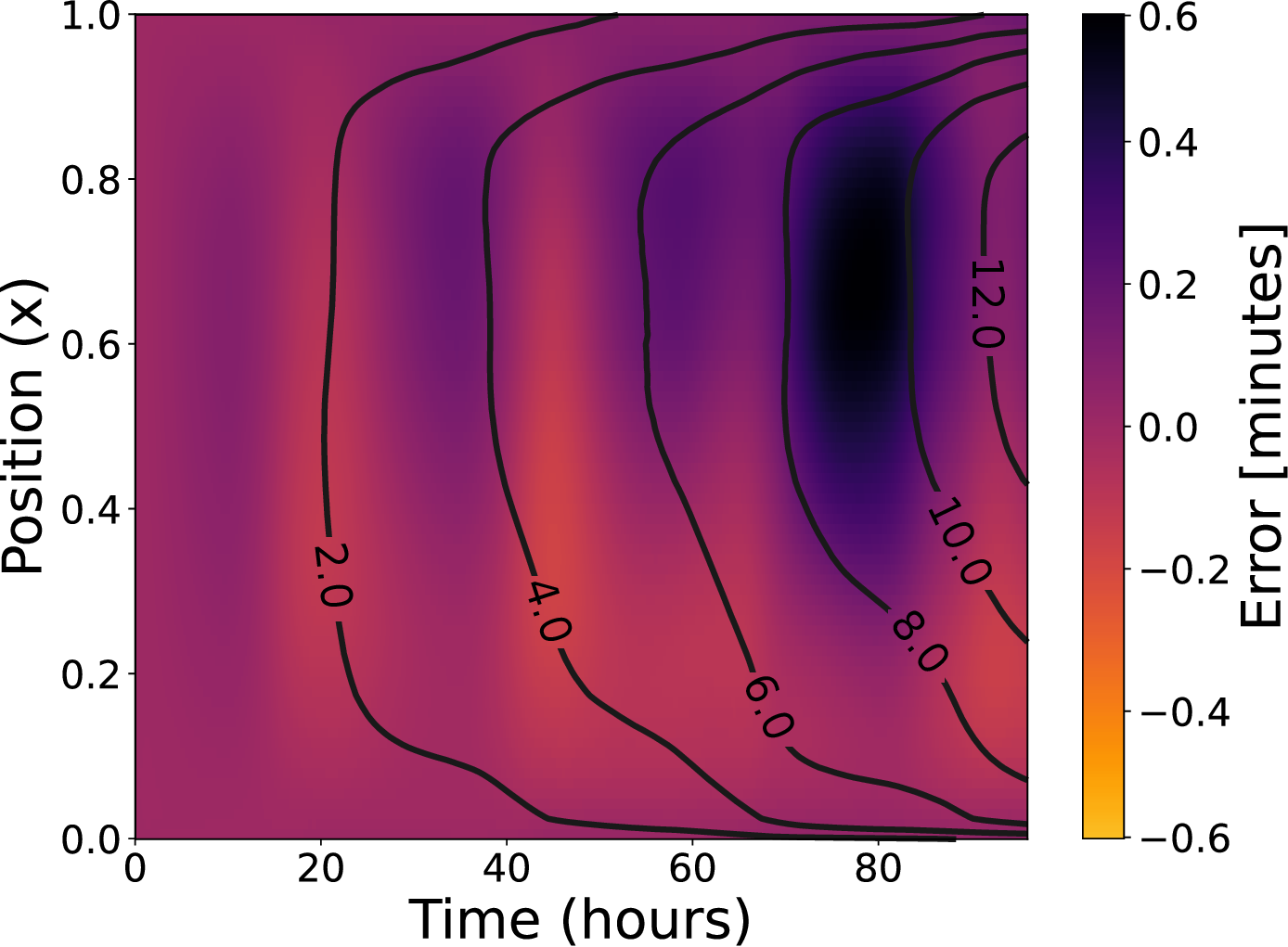 }
        \caption{}
        \label{subfig:dPINN_error}
    \end{subfigure}
    \caption{Spatio-temporal temperature prediction errors for (a) B-PINN and (b) d-PINN. The heatmap shows the mean error relative to FEM. The contour lines indicate the standard deviation of the error.}
    \label{fig:temperature_error}
\end{figure}

To further assess the accuracy and calibration of the ageing predictions, FEM-based transformer oil temperature estimates (Figure~\ref{fig:PDEmatlab}) are used as ground truth to compute the corresponding spatiotemporal insulation ageing profiles. Based on this reference, ageing prediction errors are evaluated for the B-PINN and d-PINN models with respect to the estimates shown in Figure~\ref{fig:heatmaps}. Figure~\ref{fig:temperature_error} shows the resulting mean ageing estimation error fields, with contour lines indicating the standard deviation of the error.

The results show that the variance of the ageing estimation error for the d-PINN grows substantially faster (approximately an order of magnitude faster) than that of the B-PINN. Such rapidly expanding uncertainty directly affects decision-making confidence, as excessively diffuse error bounds hinder reliable risk assessment and RUL evaluation. In contrast, the B-PINN maintains a more controlled and balanced growth of uncertainty alongside the ageing estimation error, resulting in more reliable probabilistic prognostic outputs.

\subsection{Sensitivity Analysis and Training Dynamics}
\label{ss:Ablation}

\subsubsection*{Aleatoric Uncertainty Response to Measurement Noise}
\label{ss:NumberSamples}

\begin{figure*}[!ht]
    \centering
    \begin{subfigure}[b]{0.46\textwidth}
        \centering
        \includegraphics[width=\textwidth]{images/Uncertainty_Density_BPINN_heteros_results_2.eps}
        \caption{}
        \label{subfig:a}
    \end{subfigure}
    \hspace{0.1em}
    \begin{subfigure}[b]{0.46\textwidth}
        \centering
        \includegraphics[width=\textwidth]{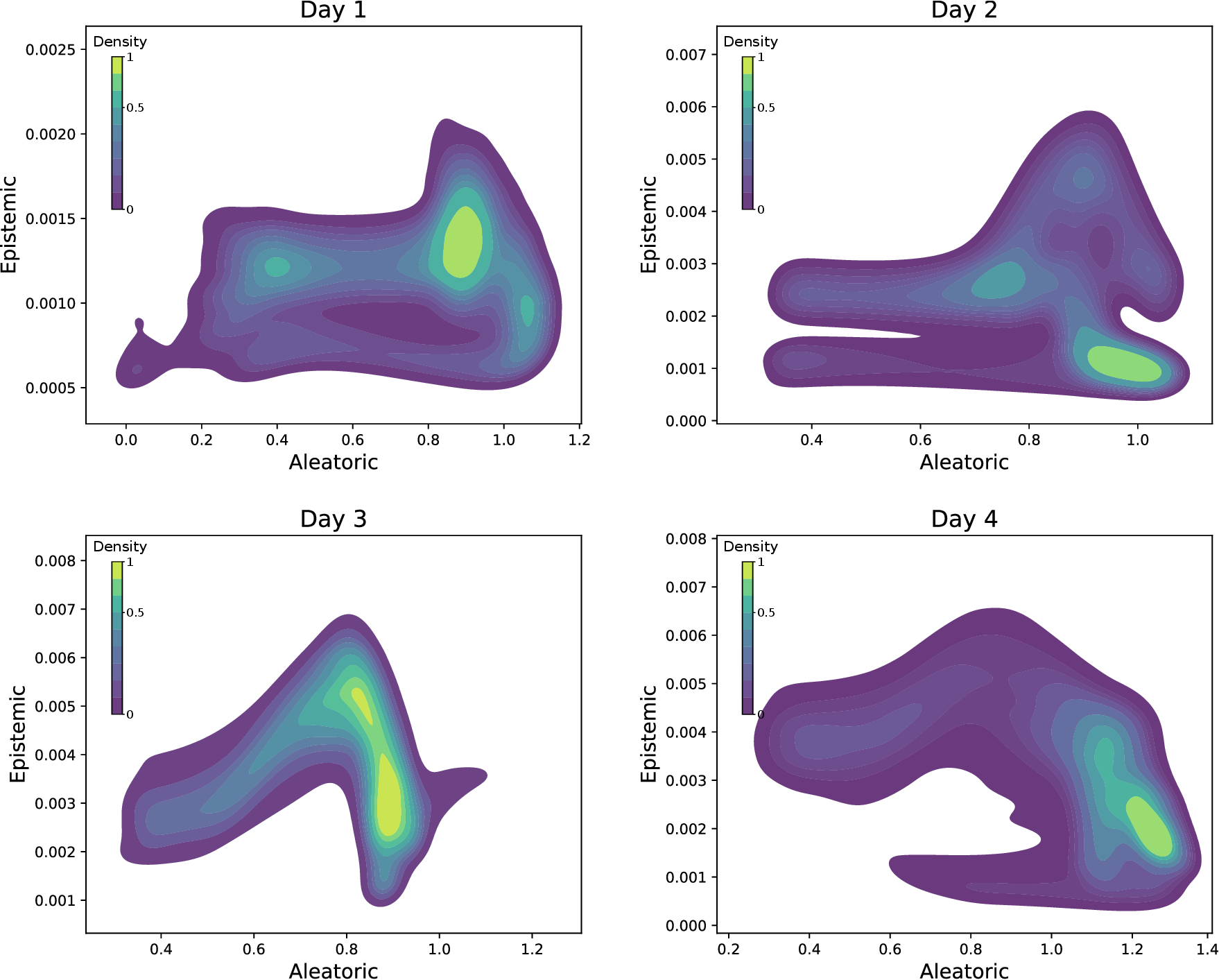}
        \caption{}
        \label{subfig:b}
    \end{subfigure}
    \begin{subfigure}[b]{0.46\textwidth}
        \centering
        \includegraphics[width=\textwidth]{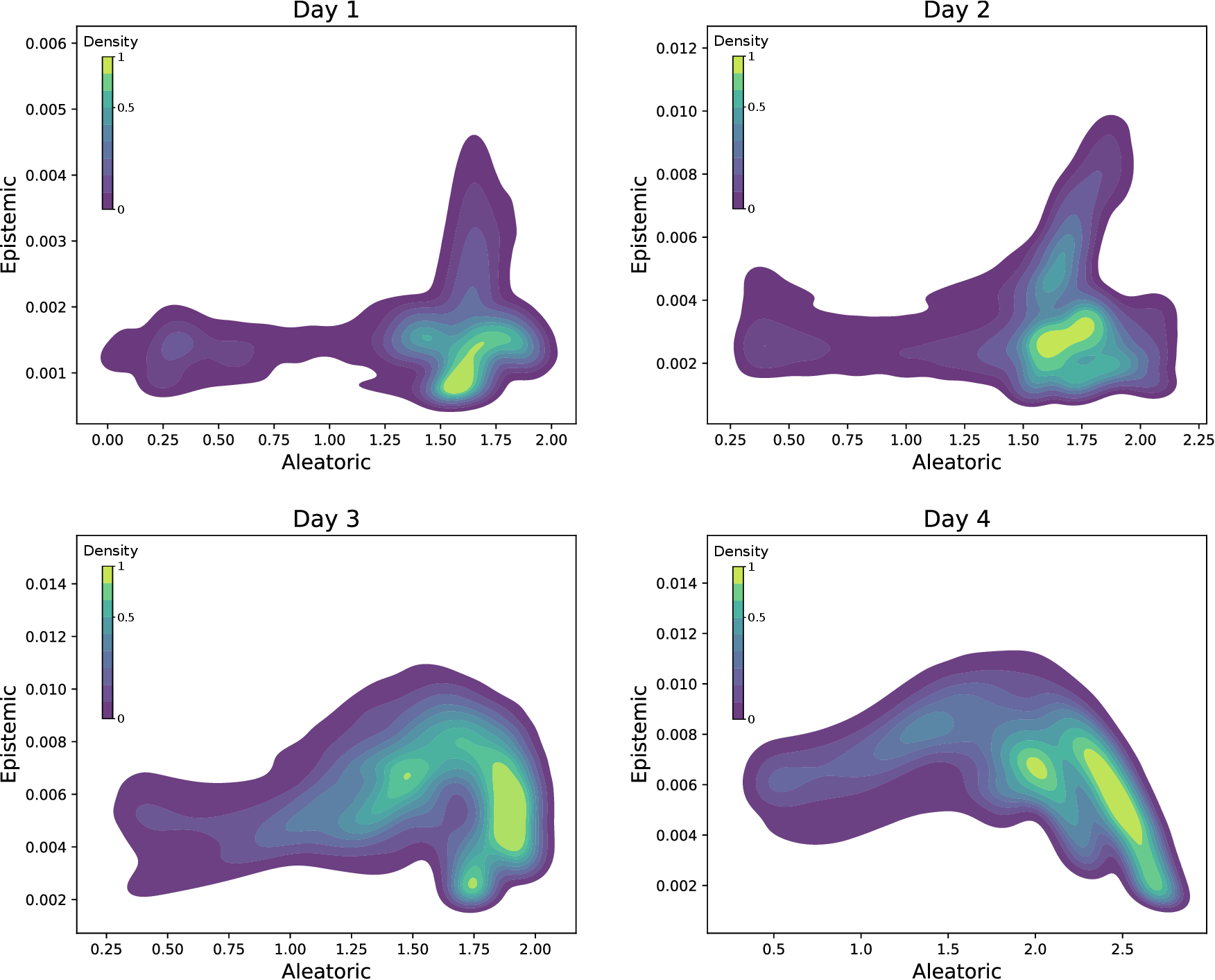}
        \caption{}
        \label{subfig:c}
    \end{subfigure}
    \hspace{0.1em}
    \begin{subfigure}[b]{0.46\textwidth}
        \centering
        \includegraphics[width=\textwidth]{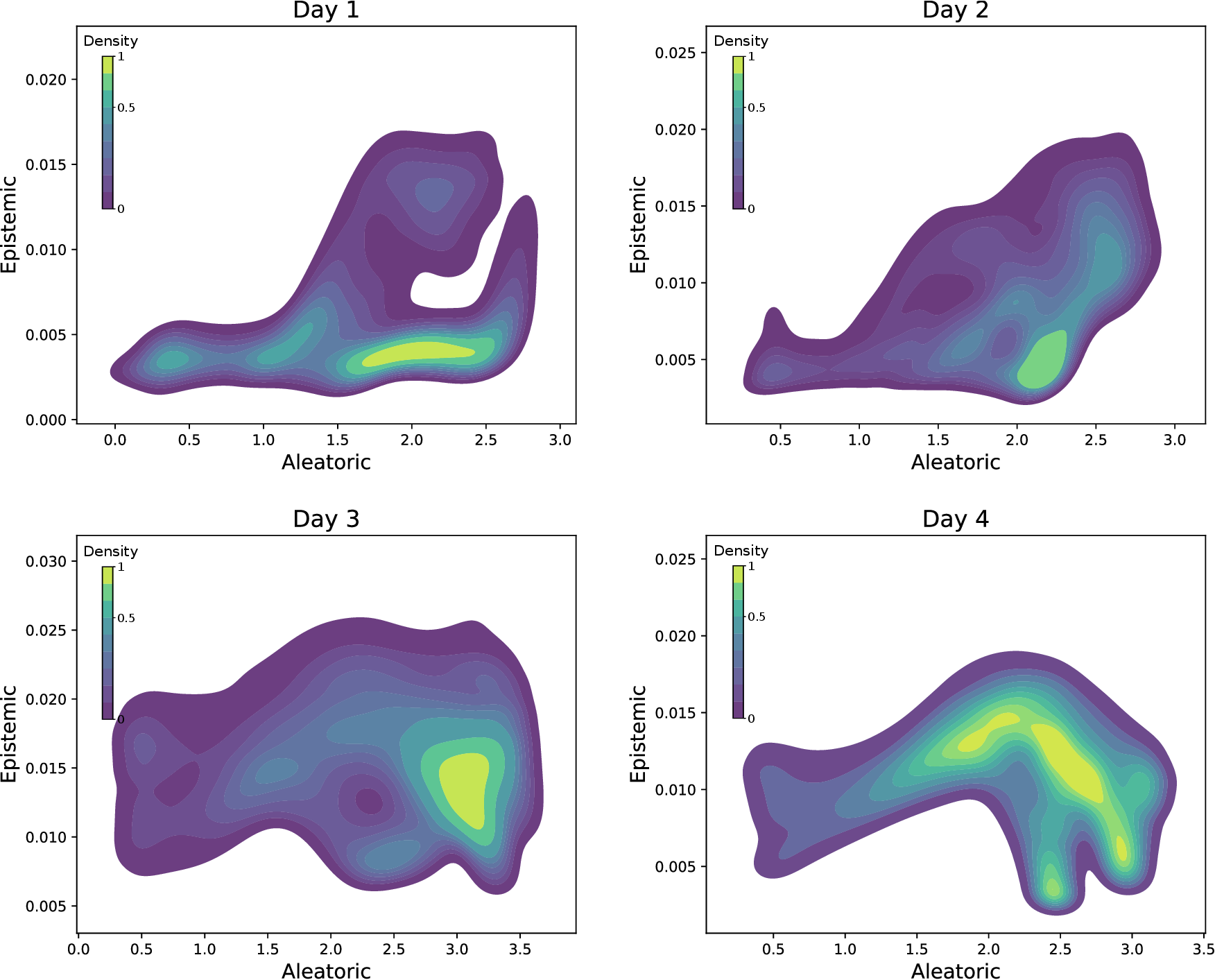}
        \caption{}
        \label{subfig:d}
    \end{subfigure}
    \caption{Comparison of joint density distributions under varying noise conditions: (a) noiseless case; (b) Gaussian noise level \%2; (c) Gaussian noise level \%4; (d) Gaussian noise level \%6.}
    \label{fig:uncertainty_density_grid}
\end{figure*}

To further evaluate the robustness of the proposed heteroscedastic B-PINN, a controlled input perturbation study was conducted by injecting Gaussian noise into the temperature and loading measurements. Measurement noise was modeled as $\epsilon \sim \mathcal{N}(0,\sigma_i), $ where $\sigma_i~=~i\times~\max(s(.,t))$, with $s(x,t)$ denoting the input signal and $i$ is the noise level in percentage units, in this case $i = \{2\%, 4\%, 6\%\}$. The perturbed signals were generated with additive noise, $s(x,t)+\sigma_i$, affecting all samples in space and time uniformly. Figure~\ref{fig:uncertainty_density_grid} shows the daily joint density distributions of aleatoric and epistemic uncertainty under increasing noise levels.

A consistent trend is observed across all analyzed days in Figure~\ref{fig:uncertainty_density_grid}. As the measurement noise increases, the joint epistemic-aleatoric distributions exhibit a systematic rightward shift along the aleatoric axis. The aleatoric uncertainty expands progressively from the noiseless case to the 6\% perturbation level, indicating that the heteroscedastic component appropriately adapts to the increased input data noise levels.

Importantly, this increase is predominantly confined to the aleatoric uncertainty part, while epistemic uncertainty remains comparatively stable in magnitude and dispersion. The joint density structure is preserved across all noise levels, without evidence of uncontrolled epistemic increase. This behavior suggests that the model maintains a separation between data-driven noise effects and model-driven uncertainty. Overall, the results demonstrate that the proposed heteroscedastic B-PINN exhibits coherent and monotonic scaling of aleatoric uncertainty in response to controlled measurement noise. This indicates proper uncertainty calibration and supports the reliability of the disentangled epistemic–aleatoric formulation under degraded sensing conditions.

\subsubsection*{Number of Initial, Residual, and Boundary Condition Samples}
\label{ss:NumberSamples}

\begin{figure}[!hb] 
\centering 
\includegraphics[width=1\linewidth]{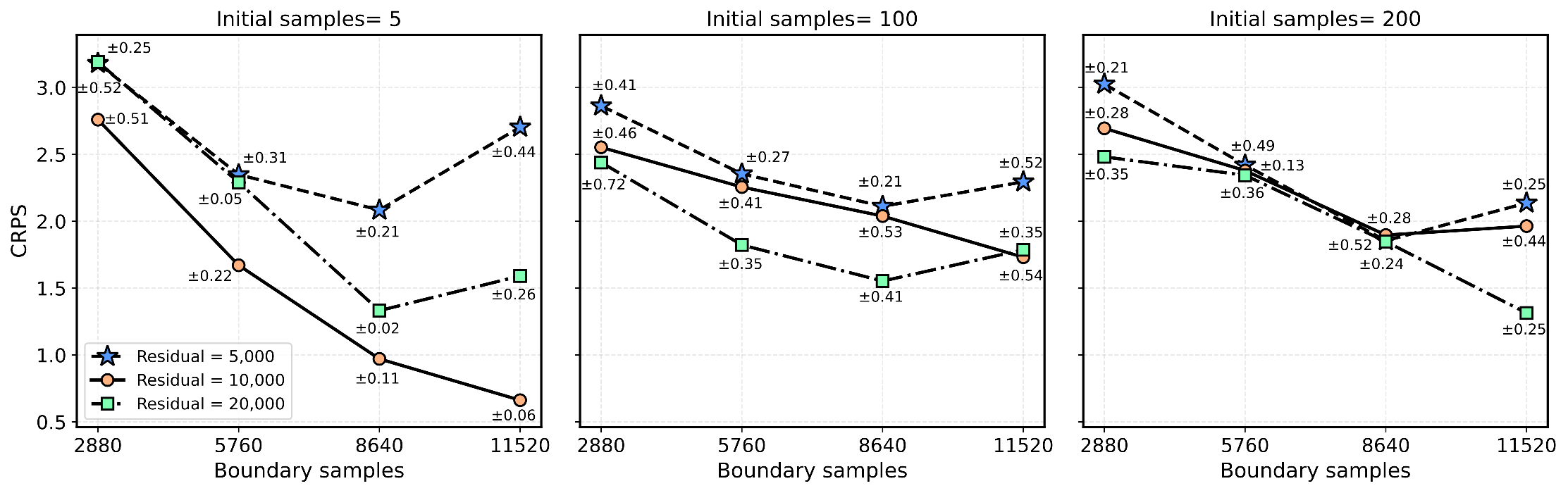} 
\caption{Sensitivity analysis of the CRPS metric with respect to the number of boundary condition samples, initial samples and residual samples. Markers indicate the mean, with annotations representing the corresponding std of the CRPS metric across runs. Lower CRPS values indicate improved UQ performance.}
\label{fig:Sensitivity_analysis} 
\end{figure}

Figure~\ref{fig:Sensitivity_analysis} shows probabilistic metrics for the heteroscedastic B-PINN approach, keeping the network architecture and prior assumptions fixed, evaluated over multiple realizations. These results summarize the impact of different sampling strategies on UQ performance by varying the number of initial condition samples ($N_i$), physics residual collocation points ($N_r$), and boundary condition samples ($N_b$). The selected configurations span sparse to moderately dense initial sampling ($N_i=\{5, 100, 200\}$), progressively increasing physics residual enforcement ($N_r=\{5000, 10000, 20000\}$), and increasing levels of boundary conditions coverage ($N_b=\{2880, 5760, 8640, 11520\}$) covering 25\%, 50\%, 75\%, and \%100 available samples for the upper and lower BCs, thereby enabling a systematic analysis of how data availability and physics constraints jointly influence probabilistic prediction quality.

Focusing on the CRPS, Figure~\ref{fig:Sensitivity_analysis} shows that probabilistic performance is primarily governed by the number of residual collocation points rather than by the number of initial condition samples. Across all configurations, the lowest CRPS values are consistently achieved with a moderate number of residual samples ($10000$), indicating an effective balance between physics enforcement and the model's ability to adapt uncertainty. Increasing the number of residual points beyond this level tends to worsen CRPS, suggesting that overly dense physics constraints can restrict the flexibility of the model and lead to less well-calibrated predictions. In contrast, increasing the number of initial samples from $5$ to $200$ does not produce systematic improvements in CRPS, indicating that the initial state is already sufficiently constrained with only a few samples. Finally, boundary condition enforcement consistently yields lowest CRPS values when all 11520 samples are used, emphasizing the stabilizing influence of boundary information on predictive accuracy and uncertainty sharpness.

The observed CRPS trends can be directly interpreted through the interplay between epistemic and aleatoric uncertainty components in the heteroscedastic B-PINN. CRPS penalizes both miscalibration and excessive dispersion, and therefore reflects the balance between epistemic and aleatoric uncertainty. When the number of residual collocation points is too small ($5000$), insufficient physics enforcement leads to increased epistemic uncertainty and broader predictive distributions, resulting in higher CRPS values. Conversely, excessively dense residual sampling ($20000$) over-constrains the solution, suppressing epistemic uncertainty, but inducing amplified aleatoric uncertainty, which again degrades CRPS.

The lowest CRPS values are consistently obtained for a moderate number of residual samples ($10000$), corresponding to a regime in which epistemic uncertainty is effectively reduced through physics-informed regularization, while retaining sufficient flexibility for input-dependent aleatoric uncertainty to be expressed. A similar interpretation applies to boundary condition sampling: full boundary enforcement ($11520$) stabilizes the solution near domain boundaries, reducing epistemic uncertainty leakage and preventing artificial increases of aleatoric uncertainty, thereby improving CRPS. In contrast, reducing boundary samples increases epistemic uncertainty near the boundaries, which propagates into wider predictive distributions and higher CRPS values.

Finally, increasing the number of initial condition samples beyond a minimal sufficient level does not systematically improve CRPS, indicating that, for the analysed problem, epistemic uncertainty associated with the initial condition is already well constrained in low $N_i$ regimes. Additional initial samples therefore contribute marginally to uncertainty reduction while potentially reducing the ability of the model to adapt aleatoric uncertainty, worsening the probabilistic performance.

\subsubsection*{Training Dynamics}
\label{ss:BPINN_Training}

\begin{figure}[!htb] 
\centering 
\includegraphics[width=0.7\linewidth]{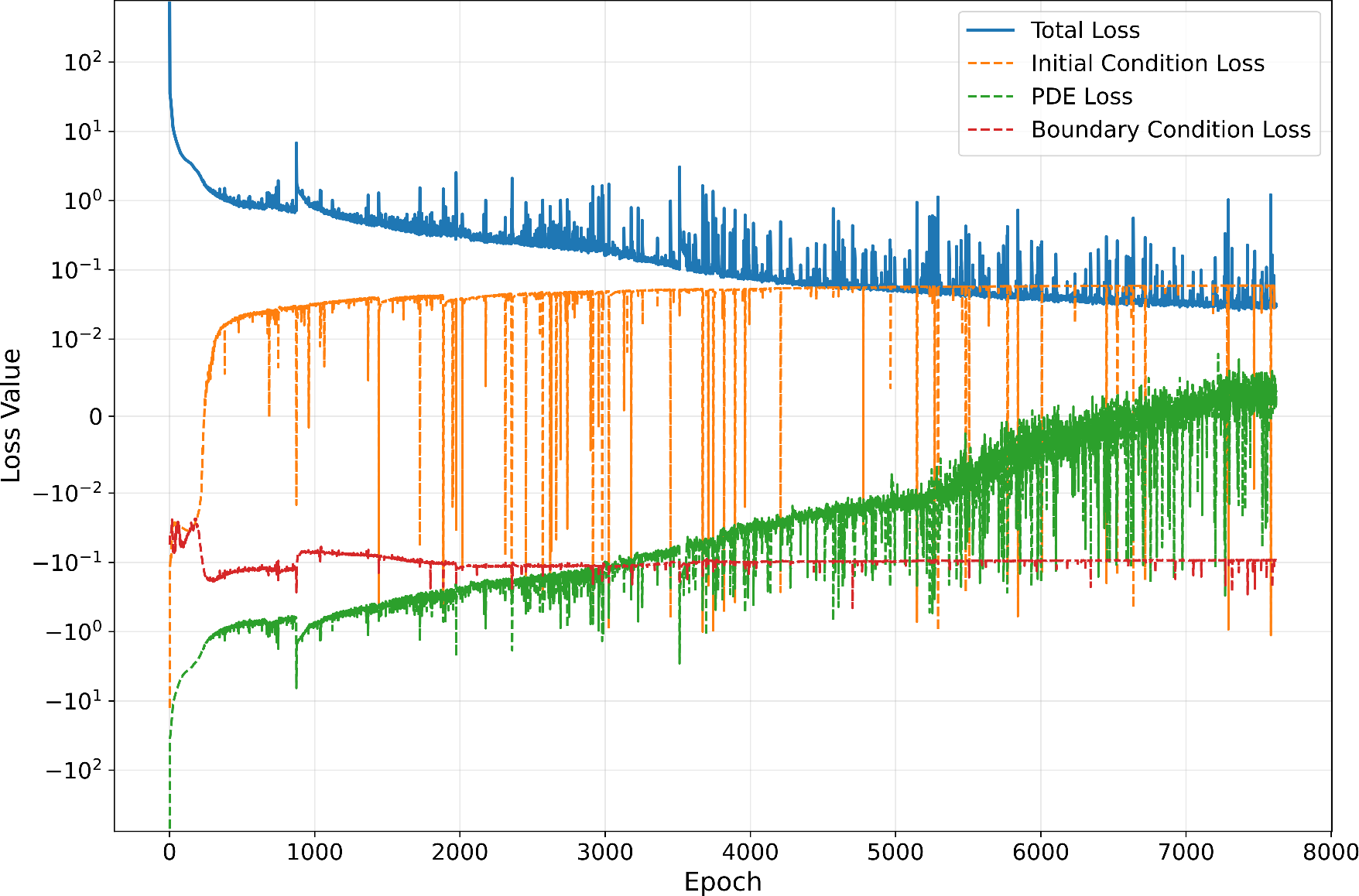} 
\caption{Evolution of the total and individual loss terms for the B-PINN configuration.}
\label{fig:BPINN_Loss} 
\end{figure}

Figure~\ref{fig:BPINN_Loss} presents the training dynamics of the heteroscedastic B-PINN. The total loss corresponds to the negative Evidence Lower Bound (negative ELBO, cf. Eq.~(\ref{eq:elbo_hetero_final})), which is composed of two main components: (i) NLL term and (ii) the KL divergence. The NLL is further decomposed into the log-likelihood contributions associated with the initial condition, boundary conditions, and PDE residual.

From Figure~\ref{fig:BPINN_Loss}, it can be observed that the initial and boundary condition terms converge rapidly during the early epochs, indicating fast alignment with the available measurement and constraint data. In contrast, the PDE residual term exhibits slower, gradual convergence, reflecting the higher complexity of satisfying the heat diffusion dynamics across the spatiotemporal domain. The KL divergence remains comparatively smooth throughout training, acting as a regularization mechanism that penalizes excessive deviation of the posterior from the prior and stabilizes optimization.

\begin{figure}[!htb] 
\centering 
\includegraphics[width=0.7\linewidth]{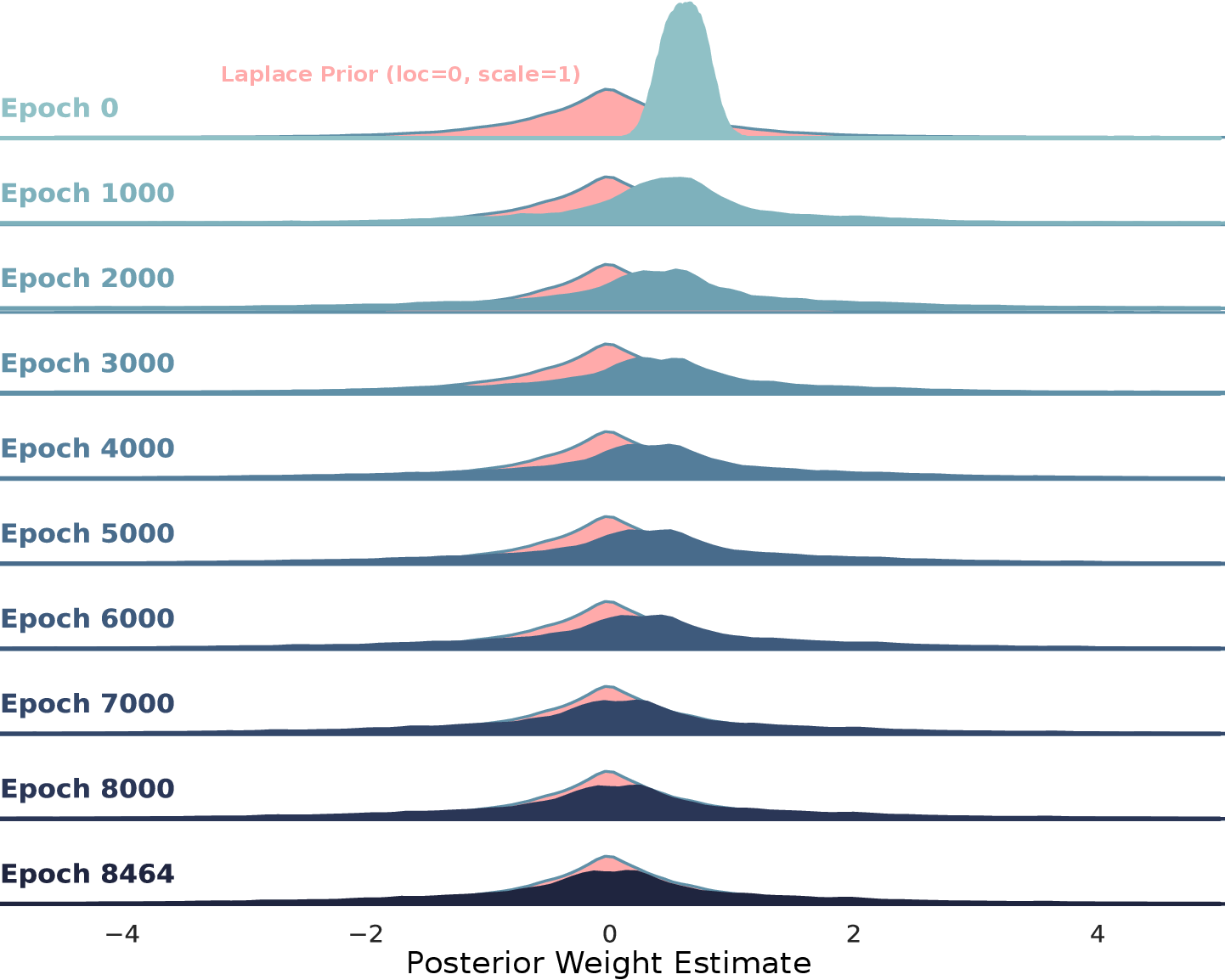} 
\caption{B-PINN posterior distribution evolution across epochs.}
\label{fig:learning_posterior} 
\end{figure}

The evolution of the posterior weight distribution is illustrated in Figure~\ref{fig:learning_posterior}. At early training stages, the posterior deviates substantially from the prior, indicating strong data-driven updates of the weight distributions. As optimization progresses, the posterior progressively stabilizes and contracts, converging towards a distribution that balances data fitting (via the log-likelihood) and complexity control (via the KL divergence). In the final epochs, the proximity between posterior and prior reflects the regularizing effect of the variational inference, preventing overfitting and enforcing controlled model complexity.

\begin{figure}[ht]
    \centering
    \begin{subfigure}[b]{0.48\textwidth}
        \includegraphics[width=\textwidth]{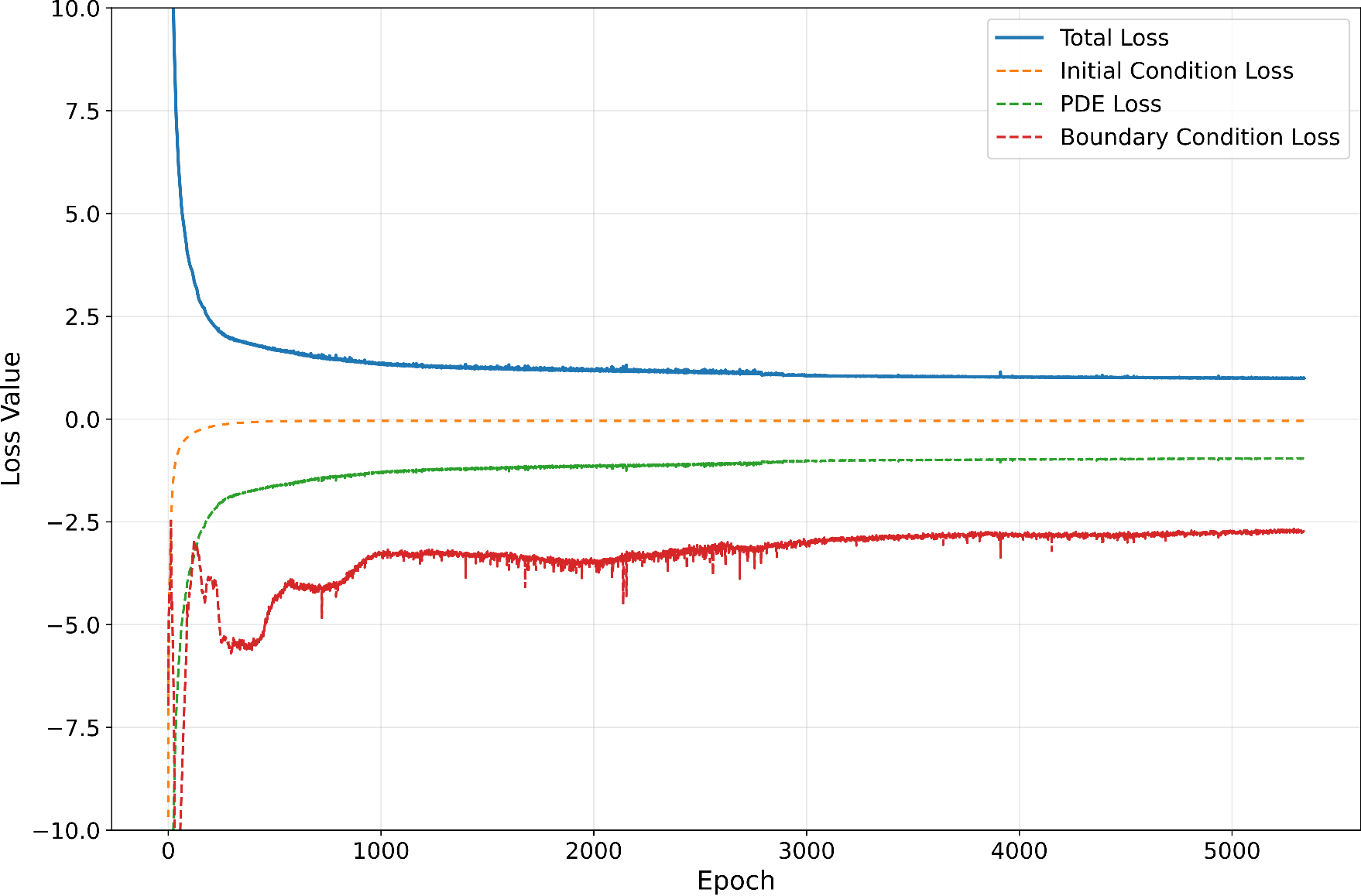}
        \caption{}
        \label{subfig:learning_dpinn}
    \end{subfigure}
    \begin{subfigure}[b]{0.48\textwidth}
        \includegraphics[width=\textwidth]{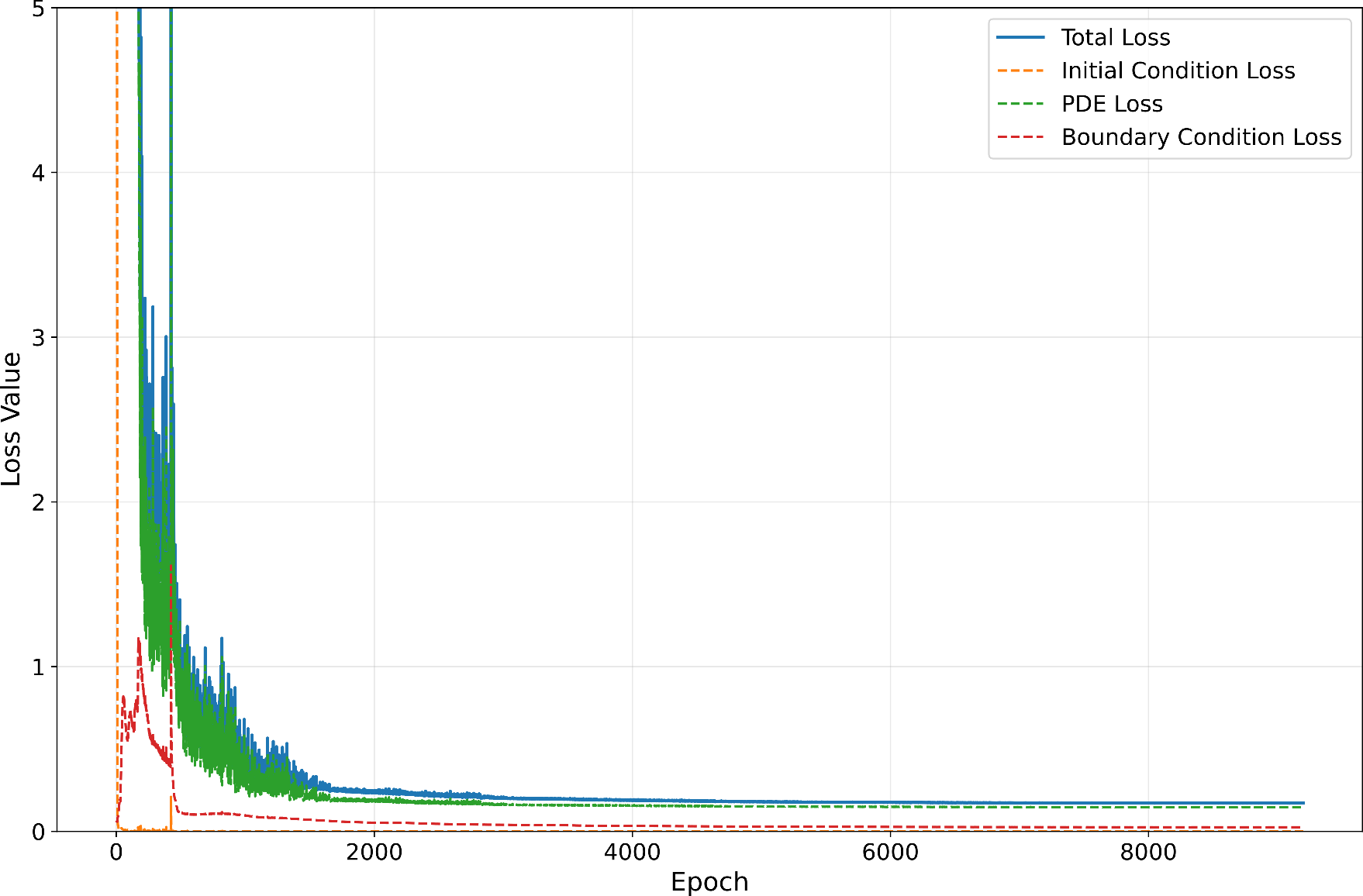}
        \caption{}
        \label{subfig:learning_pinn}
    \end{subfigure}
    \caption{Evolution of the training loss for (a) d-PINN and (b) vanilla PINN configurations.}
    \label{fig:PINN_DPINN_Training}
\end{figure}

Figure~\ref{fig:PINN_DPINN_Training} compares the convergence behavior of the heteroscedastic d-PINN and the vanilla PINN configurations. Early stopping criteria were applied in all cases, which results in differences in total training epochs across configurations.

It can be observed in Figure~\ref{fig:PINN_DPINN_Training} that the d-PINN exhibits similar convergence dynamics to the B-PINN in terms of loss decomposition, as it is trained through a likelihood-based objective. However, unlike the B-PINN, it does not include an explicit KL regularization term. The vanilla PINN, trained using a mean-squared-error formulation for the data terms combined with the PDE residual loss, shows a monotonic decrease of both residual and total losses, with rapid convergence of the initial and boundary condition terms.

The PINN loss function includes different terms that must be optimized. Each individual term goes through a minimization process ruled by the optimization algorithm and the loss function. However, each term is specified in a different information space, \textit{i.e.} residual loss terms operate with physics-residuals and data-loss terms directly operate with error residuals. The slowest convergence rate among loss terms, generally dominates the overall solution convergence \cite{Chen25}. There have been presented different loss term weight adaptation methods for PINNs, such as Residual Based Attention \cite{Ramirez_25}, Neural Tangent Kernel based methods \cite{Wang2022}, or Self Adaptive methods \cite{Chen25}.

In contrast, the B-PINN objective is entirely formulated in the probabilistic space. Therefore, the adoption of loss term weight adaptation mechanisms proposed for PINNs are not applicable. For a direct comparative among tested configurations, no weight adaptation mechanisms have been implemented and all loss terms have been weighted manually. The implementation of loss term weight adaptation mechanisms for B-PINNs is an area of research that will be addressed in future work.

\section{Conclusion}
\label{s:Conclusions}

Scientific Machine Learning (SciML) integrates physical knowledge with machine learning (ML), yielding predictive models that generalize more robustly than purely data-driven approaches while requiring less data. Hybrid Prognostics and Health Management (PHM) methods share conceptual similarities with SciML, as both combine physics-of-failure models with data-driven components. However, SciML provides a more principled learning framework in which physical constraints are embedded directly into the training process. To the best of the authors’ knowledge, this paradigm has not yet been fully explored within PHM in general, and for insulation material prognostics tasks in particular.

This work proposes a heteroscedastic Bayesian Physics-Informed Neural Network (B-PINN) framework for probabilistic insulation ageing estimation in electrical transformers, explicitly capturing epistemic and aleatoric sources of uncertainty. The proposed approach embeds the heat diffusion partial differential equation into the learning process while leveraging operational measurements of transformer loading and temperature. Epistemic uncertainty is represented by modeling Neural Network weights probabilistically, whereas heteroscedastic aleatoric uncertainty is learned through an input-dependent noise model enabled by architectural and training modifications to the B-PINN formulation.

Numerical experiments demonstrate that the proposed heteroscedastic B-PINNs consistently outperforms its homoscedastic counterpart across the evaluated configurations, yielding improved predictive accuracy, better-calibrated uncertainty estimates, and enhanced reliability. In terms of sharpness, the heteroscedastic B-PINN produces, on average, sharper predictive distributions, albeit with increased variability associated with localized adaptations to input-dependent uncertainty. In contrast, heteroscedastic dropout PINNs (d-PINNs) exhibit degraded performance, characterized by reduced predictive accuracy and systematically over-dispersed uncertainty estimates, particularly at long prediction instants.

A detailed decomposition of aleatoric and epistemic uncertainty in the heteroscedastic B-PINN configuration shows that most uncertainty realizations remain concentrated within narrow aleatoric–epistemic regimes, indicating stable uncertainty behavior. Occasional extended tails correspond to transient periods of data scarcity or localized model mismatch.

From a prognostics perspective, the resulting ageing estimates confirm that the heteroscedastic B-PINN provides the most reliable and informative predictions. Conversely, the heteroscedastic d-PINN yields broad uncertainty bounds that hinder maintenance decision-making and may lead to overly conservative asset replacement strategies.

A sensitivity analysis of the heteroscedastic B-PINN with respect to the number of initial condition, boundary condition, and residual collocation points, further provides practical insights into the configuration of B-PINNs for prognostics applications. For the analyzed case study: (i) a moderate number of residual collocation points yields optimal Continuous Ranked Probability Score (CRPS) values, reflecting a balanced trade-off between physics enforcement and uncertainty adaptability; (ii) the influence of initial condition samples is limited when the initial state is sufficiently constrained; and (iii) full boundary condition enforcement leads to the best CRPS performance, highlighting the stabilizing role of boundary information in both predictive accuracy and uncertainty sharpness.

Beyond methodological contributions, the proposed heteroscedastic B-PINN framework offers tangible benefits for transformer asset management. By explicitly quantifying and propagating epistemic and aleatoric uncertainty, the approach supports informed decision-making under data scarcity and localized model mismatch, mitigating the risks associated with overly conservative or unsafe maintenance strategies.

The present implementation relies on variational inference for posterior approximation. Future work will explore alternative Bayesian inference techniques and physics-informed priors, to further enhance uncertainty quantification and robustness in SciML-based PHM frameworks.

\medskip
\noindent \textbf{Acknowledgements} \par 
This work has been partially funded by the Spanish State Research Agency (grant No. PID2024-156284OA-I00) and the Basque Government (grants IT1504-22, KK-2026/00055). Jose I. Aizpurua is funded by the Ramón y Cajal Fellowship, Spanish State Research Agency (grant No. RYC2022-037300-I), co-funded by MCIU/AEI/10.13039/501100011033 and FSE+.

\medskip
\noindent \textbf{Conflict of Interest} \par 
The authors declare no conflict of interest.
\medskip

%
\bibliographystyle{ieeetr} 
\bibliography{Biblio}

\appendix

\section{Transformer Thermal Model}
\label{s:Trafo}

Transformers are key assets for the reliable operation of the power grid. The increasing penetration of renewable energy sources to the grid affects the transformer's health \cite{Aizpurua_23}. The main insulating material for oil-immersed transformers is paper immersed oil, and their main failure mode is the insulation degradation. The insulating paper is made of cellulose polymer and the degree of polymerization determines the strength of the insulating paper \cite{IEC60076_transf12}. The insulation paper degradation is directly caused by the thermal stress. 

Accordingly, this section reviews the main transformer thermal modelling steps, including the oil temperature estimation stage (Section~\ref{ss:PDE}) and subsequent winding temperature estimation stage, which is used to calculate the hottest-spot temperature (HST), \textit{i.e.} highest insulation temperature (Section~\ref{ss:Winding}). Finally, the thermal model is connected to the insulation aging assessment model, which is used to estimate the loss of life of the insulation (Section~\ref{ss:AgeingEstimate}).

\subsection{Spatiotemporal Oil Temperature Modelling}
\label{ss:PDE}

The spatial distribution of oil and winding temperature is key for the cost-effective transformer health management. To capture this, a thermal modelling approach is developed  based on partial differential equations (PDE). The heat diffusion PDE is considered to model the temporal and spatial evolution of the transformer oil temperature ($\Theta_{O}(x,t)$).  Due to the transformer oil characteristics, radiative heat diffusion is considered and not convection. The general form of the one-dimensional heat diffusion equation is defined as \cite{Ramirez_25}:

\begin{equation}
	\label{eq:PDE_1D_Diffusion}
	\begin{split}
		k\frac{\partial^2\Theta_{O}(x,t)}{\partial x^2} + q(x,t) &= \rho c_p \frac{\partial \Theta_{O}(x,t)}{\partial t} \\
		\frac{\partial^2\Theta_{O}(x,t)}{\partial x^2} + \frac{1}{k}q(x,t) &= \frac{1}{\alpha} \frac{\partial \Theta_{O}(x,t)}{\partial t}
	\end{split}
\end{equation}

\noindent where $x,t\in\mathbb{R}$ are the independent variables, which denote position [m] and time [s], respectively, $\Theta_{O}(x,t)$ is given in Kelvin [K], $k$ is the thermal conductivity [W/m.K], $c_p$ is the specific heat capacity [J/kg.K], $\rho$ is the density [kg/m\textsuperscript{3}], $q(x,t)$ is the rate of heat generation [W/m\textsuperscript{3}], and $\alpha=\frac{k}{\rho c_p}$ is the thermal diffusivity [m\textsuperscript{2}/s].

\begin{figure}[!htb]
	\centering
	\includegraphics[width=.45\columnwidth]{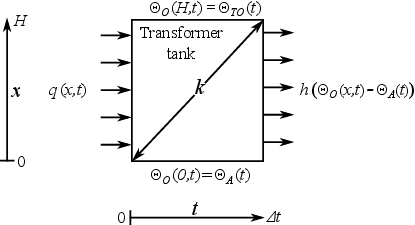}
	\caption{Transformer heat diffusion model \cite{Ramirez_25}.}
	\label{fig:Trafo_Diffusion_Basic}
\end{figure}

Figure~\ref{fig:Trafo_Diffusion_Basic} shows the thermal parameters of the transformer of the heat diffusion model, which considers the heat source, $q(x,t)$, and the convective heat transfer, $h(\Theta_{O}(x,t)-\Theta_{A}(t))$.

The evolution of the heat source in space and time, $q (x,t)$, is defined as follows:

\begin{equation}
	\label{eq:heat-source}
	q(x,t)=P_0+P_K(t)-h\left(\Theta_{O}(x,t)-\Theta_{A}(t)\right)
\end{equation}

\noindent where $\Theta_{A}(t)$ is the ambient temperature, $h$ is the convective heat transfer coefficient, $P_0$ is the no load losses [W], and $P_K(t)$ is the load losses, defined as follows:

\begin{equation}
	P_K(t)=K(t)^2\mu,
\end{equation}

\noindent where $K(t)$ is the load factor [p.u.], and $\mu$ is the rated load losses [W].

The challenge is to accurately model the transformer oil temperature vertically along its height $H$ (Figure~\ref{fig:Trafo_Diffusion_Basic}). It can be observed that the spatial distribution is considered along the vertical axis ($x$). It is assumed that $\Theta_{O}(x,t)$ is equal to $\Theta_{A}(t)$ at the bottom ($x$\hspace{0.5mm}=\hspace{0.5mm}$0$) and $\Theta_{TO}(t)$ at the top ($x$\hspace{0.5mm}=\hspace{0.5mm}$H$). Namely, these are the Dirichlet boundary conditions of the PDE that is going to be solved:

\begin{equation}
	\label{eq:BoundaryConditions}
	\begin{split}
		\Theta_{O}(0,t)&=\Theta_{A}(t)\\
		\Theta_{O}(H,t)&=\Theta_{TO}(t)\\
	\end{split}
\end{equation}

\subsection{Hotspot Temperature (HST) Modelling}
\label{ss:Winding}

The HST is the most critical thermal indicator. However, direct measurements are difficult and expensive. Generally the HST value, $\Theta_{H}(t)$, is estimated indirectly from the top-oil temperature (TOT), $\Theta_{TO}(t)$, defined as \cite{IEC60076_transf12}:

\begin{equation}
	\label{eq:HST}
	\Theta_{H}(t)=\Theta_{TO}(t)+\Delta\Theta_{H}(t)
\end{equation}

\noindent where $\Delta\Theta_{H}(t)$ is HST rise over TOT and $t$\hspace{0.4mm}$\in$\hspace{0.4mm}$\mathbb{R}$ is time.

With the spatiotemporal oil temperature estimate $\hat{\Theta}_{O}(x,t)$, the winding temperature distribution, $\hat{\Theta}_{W}(x,t)$, is defined as follows:

\begin{equation}
	\label{eq:HST_spatial}
	\hat{\Theta}_{W}(x,t)={\hat{\Theta}_{O}}(x,t)+\Delta\Theta_{H}(t)
\end{equation}

\noindent where, $x,t$\hspace{1.0 mm}$\in$\hspace{0.4mm}$\mathbb{R}$ are the position and time, respectively, $\hat{\Theta}_{O}(x,t)$ is the spatiotemporal oil temperature estimate. The hottest spatial temperature in Eq.~(\ref{eq:HST_spatial}) at each time instant $t$, corresponds to the HST in Eq.~(\ref{eq:HST}), \textit{i.e.} $max(\hat{\Theta}_{W}(\cdot,t))$=$\Theta_{H}(t)$.

$\Delta\Theta_{H}(t)$ is the HST rise over TOT, which is given by:

\begin{equation}
	\label{eq:delta_HST}
	\Delta\Theta_{H}(t)=\Delta\Theta_{H_1}(t)-\Delta\Theta_{H_2}(t)
\end{equation}

\noindent where $\Delta\Theta_{H_1}(t)$ and $\Delta\Theta_{H_2}(t)$ model the oil heating considering the HST variations defined as follows \cite{IEC60076_transf12}:

\begin{equation}
	\label{eq:HST_Transient1_1}
	d\Delta\Theta_{H_i}(t)=\upsilon_i\left[\beta_iK(t)^y-\Delta\Theta_{H_i}(t)\right]
\end{equation}

\noindent where {$K(t)$} is the load factor {[p.u.]}, $y$ is the winding exponent constant, which models the loading exponential power with the heating of the windings, {$i$\hspace{0.3mm}=\hspace{0.3mm}\{1,\hspace{0.4mm}2\}}, {$\upsilon_1$\hspace{0.2mm}=\hspace{0.3mm}$\Delta t/k_{22}\tau_w$}, {$\beta_1$\hspace{0.2mm}=\hspace{0.3mm}{$k_{21}\Delta\Theta_{H,R}$}} both for {$i$\hspace{0.3mm}=\hspace{0.3mm}$1$}, and {$\upsilon_2$\hspace{0.2mm}=\hspace{0.3mm}{$k_{22}\Delta t/\tau_{TO}$}}, {$\beta_2$\hspace{0.2mm}=\hspace{0.3mm}$(k_{21}$\hspace{0.3mm}-\hspace{0.3mm}$1)${$\Delta\Theta_{H,R}$}} {both} for {$i$\hspace{0.3mm}=\hspace{0.3mm}$2$}. {$\Delta t$}\hspace{0.3mm}{=}\hspace{0.3mm}$t$\hspace{0.5mm}-\hspace{0.5mm}$t${$^\prime$}, {$\tau_w$} and {$\tau_{TO}$} are the winding and oil time constants, {$k_{21}$} and {$k_{22}$} are the transformer thermal constants, and {$\Delta\Theta_{H,R}$} is the HST rise at rated load. The operator {$d$} denotes a difference operation on $\Delta t$, such that {$d\Delta\Theta_{H_i}(t)$\hspace{0.3mm}=\hspace{0.3mm}$\Delta\Theta_{H_i}(t)$\hspace{0.3mm}-\hspace{0.3mm}$\Delta\Theta_{H_i}(t^\prime)$} also for {$i$\hspace{0.3mm}=\hspace{0.3mm}\{1,\hspace{0.4mm}2\}}. To guarantee numerical stability, {$\Delta t$} should be small, never greater than half of the smaller time constant. 

Under steady state, the initial condition, {$\Theta_{H}(0)$}, can be defined as:

\begin{equation}
	\label{eq:InitCond_HST}
	\Theta_{H}(0)\!=\!\Theta_{TO}(0)\!+\!k_{21}\Delta\Theta_{H,R}K(0)^y\!-\!(k_{21}\!-\!1)\Delta\Theta_{H,R}K(0)^y
\end{equation}

Eq.~(\ref{eq:InitCond_HST}) allows iteratively estimating the next HST values, $\Theta_{H}(n\Delta t)$, $n\in\mathbb{Z}^+$, using Eqs.~(\ref{eq:HST}), (\ref{eq:delta_HST}), and (\ref{eq:HST_Transient1_1}).

\subsection{Ageing Assessment}
\label{ss:AgeingEstimate}

The IEC 60076-7 standard defines insulation ageing acceleration factor at time $t$, {$V(t)$}, as \cite{IEC60076_transf12}:

\begin{equation}
	\label{eq:ageing_factor}
	V(t)=2^{\left(\Theta_{H}(t)-98\right)/6}
\end{equation}

Using the spatiotemporal winding temperature $\hat{\Theta}_{W}(x,t)$, the ageing acceleration factor at time $t$ and position $x$, {$V(x,t)$}, can be defined as:

\begin{equation}
	\label{eq:ageing_factor_spatial}
	V(x,t)=2^{\left(\hat\Theta_W(x,t)-98\right)/6}
\end{equation}

The IEC 60076-7 assumes an expected life of 30 years, with a reference HST of 98$^\circ$C \cite{IEC60076_transf12}. The loss of life (LOL) at location $x$ and time $t$ can be defined as:

\begin{equation}
	\label{eq:LoL}
	LOL(x,t)=\int_0^t V(x,t)dt
\end{equation}

Consequently, the LOL at discrete time $L\Delta t$ and location $x$ can be obtained by summing the ageing (cf. Eq.~(\ref{eq:ageing_factor_spatial})) evaluated at the same time instants:

\begin{equation}
	\label{eq:lifetime_spatial}
	LOL(x,L\Delta t)=\sum_{n=0}^{L} V(x,n\Delta t)
\end{equation}

\noindent where  $n,L$\hspace{1.0 mm}$\in$\hspace{0.4mm}$\mathbb{Z}$\hspace{0.1mm}$^+$.

LOL can be converted into a recurrence relation for remaining useful life estimation \cite{Aizpurua_23}.

\subsection{Transformer Parameters}
\label{ss:TrafoDetails}

The transformer's nameplate parameters are summarized in Table~\ref{table:CaseStudy_trafo}.

\begin{table}[!hbtp]
	\centering
	\caption{Transformer parameter values.}
	\label{table:CaseStudy_trafo}
	\begin{tabular}{|c|c|}
		\hline
		{\textbf{Parameter}} & {\textbf{Value}}\\ \hline	
		{Rating [kVA], V\textsubscript{1}/V\textsubscript{2}} & {1100, 22000/400}\\ 
		{R=Load losses/No load losses [W]} & {9800/842}\\ 		
		{$\Delta\Theta_{H,R}$} [$^\circ$C] & {15.1}\\ 	
		{$k_{21}$, $k_{22}$} & {2.32, 2.05}\\ 	
		{$\tau_0$, $\tau_w$ [min.]} & {266.8, 9.75} \\ 	\hline	
	\end{tabular}
\end{table}

\section{Benchmarking Models and Metrics}
\label{s:benchmarking_metrics}

The proposed B-PINN approach is benchmarked against (i) a dropout based PINN implementation, which has been selected due to its computational efficiency \cite{Zhang_19} and (ii) the B-PINN homoscedastic counterpart which models epistemic and constant aleatoric sources of uncertainty \cite{BPINN_PHM25}.

\subsection{Dropout-PINN}
\label{ss:DropoutPINN}

\begin{figure}[!ht]
    \centering
    \includegraphics[width=0.75\textwidth]{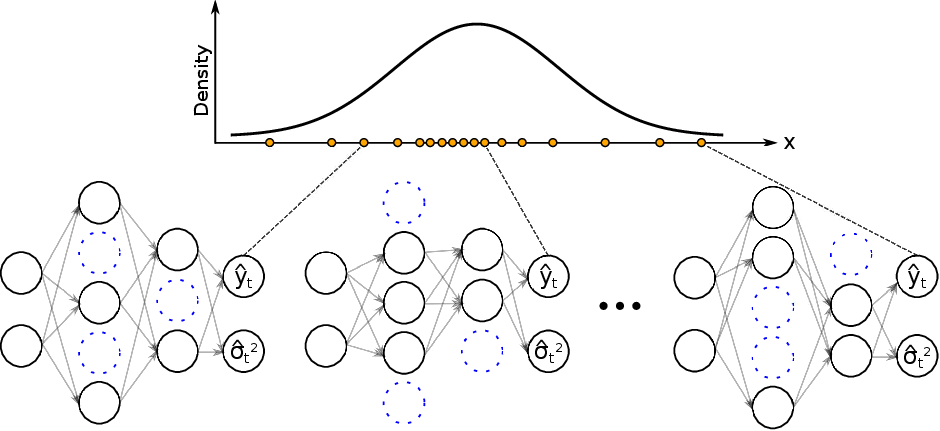}
    \caption{Use of dropout at inference stage with MC sampling.}
    \label{fig:mc_dropout}
\end{figure}

Dropout-PINN (d-PINN) extends the vanilla PINN (Section~\ref{ss:PINN_Basics}) by inserting dropout layers after each hidden layer, with a dropout rate, which is selected through a hyperparameter tuning process (introduced in Table~\ref{tab:dropout_pinn_hyperparameters}). During training, dropout randomly deactivates a subset of neurons, acting as a regularization mechanism that improves generalization and mitigates overfitting~\cite{Gal16}. Unlike standard inference, dropout is kept active at inference time. As illustrated in Fig.~\ref{fig:mc_dropout}, $M$ stochastic forward passes are performed during testing, each corresponding to a different realization of the dropout mask. This Monte Carlo sampling procedure yields an ensemble of network predictions whose empirical distribution enables the estimation of epistemic uncertainty through the computation of the predictive variance~\cite{Zhang_19}.

Finally, in order to capture the aleatoric uncertainty, and assuming that the uncertainty can be modelled with an equivalent Gaussian distribution, $N(\mu,\sigma)$, the d-PINN architecture has been modified adding an additional output which models variance, $\sigma^2$. By modifying the architecture and employing a loss function based on NLL, the uncertainty can be learned as a function of data~\cite{kendall2017}:

\begin{equation}
\label{eq:NegLogLiklossfuntion}
\mathcal{L}_{NN}(y_{true}|\mu,\sigma^2) = \frac{1}{N} \sum_{i=1}^{N} \frac{(y_i - \mu(x_i))^2 }{2\sigma (x_i)^2} + \frac{1}{2}log\sigma(x_i)^2 
\end{equation}

\noindent where $y_i$ is the true value for the $i$-th element in the batch, $\mu(x_i)$ is the predicted mean value for the $i$-th element, and $\sigma(x_i)^2$ represents the variance.

The hyperparameters defined for the d-PINN model (Table~\ref{tab:dropout_pinn_hyperparameters}) were tuned using the governing 1D heat diffusion partial differential equation (see Appendix~\ref{s:Trafo}). The best results were obtained using the Adam optimizer with a learning rate of 0.01, a hyperbolic tangent activation function, and a fully connected architecture consisting of [2, 50, 50, 2] neurons.

\begin{table}[h!]
    \centering
    \setlength{\tabcolsep}{3pt}
    \caption{Architecture and optimization hyperparameters for d-PINN along with grid search range and final values.}
    \begin{tabular}{llll}
        \toprule
        \textbf{Category} & \textbf{Hyperparameter} & \textbf{Grid} & \textbf{Value} \\
        \midrule
        \multirow{4}{*}{Architecture}
        & \#Layers & 2,3,4 & 2 \\
        & \#Nodes & 16,32,50,64 & 50 \\
        & Activation & ReLU,Tanh & Tanh \\
        & Dropout rate & 0.05,0.1,0.2,0.3 & 0.1 \\
        \midrule
        \multirow{6}{*}{Optimization}
        & Loss Function & - & -ELBO \\
        & Optimizer & - & Adam \\
        & Learning Rate & 0.001,0.01,0.1 & 0.01 \\
        & Batch size & 8,16,32 & 16 \\
        & \#Epochs & 5000,10000,15000 & 15000 \\
        & \#MC inferences & 50,100,200 & 200 \\
        \bottomrule
    \end{tabular}
    \label{tab:dropout_pinn_hyperparameters}
\end{table}

The dropout rate strongly affects the quality of the estimated uncertainty. Consequently, a sensitivity analysis is conducted to identify the dropout rate that provides the most reliable uncertainty estimates, following the procedures in \cite{Alcibar_2025} and \cite{BPINN_PHM25}. Table~\ref{table:Dropout_PINN} displays the obtained results.

\begin{table}[!htb]
    \centering
    \setlength{\tabcolsep}{5pt}
    \caption{Sensitivity of the dropout rate in dropout-PINN. Best results highlighted in boldface.}
    \begin{tabular}{clll}
        \toprule
        \textbf{Dropout Rate} & \textbf{RMSE ($\downarrow$)} & \textbf{CRPS ($\downarrow$)} & \textbf{NLL($\downarrow$)}  \\
        \midrule
        \textbf{0.05} & \textbf{2.051}  &  \textbf{1.205}  &  \textbf{2.219}  \\ 
        0.10 & 2.091  &  1.207  &  2.289  \\ 
        0.15 & 2.685  &  1.484  &  2.341  \\ 
        0.20 & 3.228  &  1.744  &  2.475  \\ 
        0.25 & 3.723 &  1.801  &  2.505  \\ 
        \bottomrule
    \end{tabular}
    \label{table:Dropout_PINN}
\end{table}

\newpage

\subsection{Bayesian PINN: Epistemic and Homoscedastic Uncertainty Quantification}
\label{s:BPINN_PHM}

\begin{figure}[!htb]
	\centering
	\includegraphics[width=0.95\linewidth]{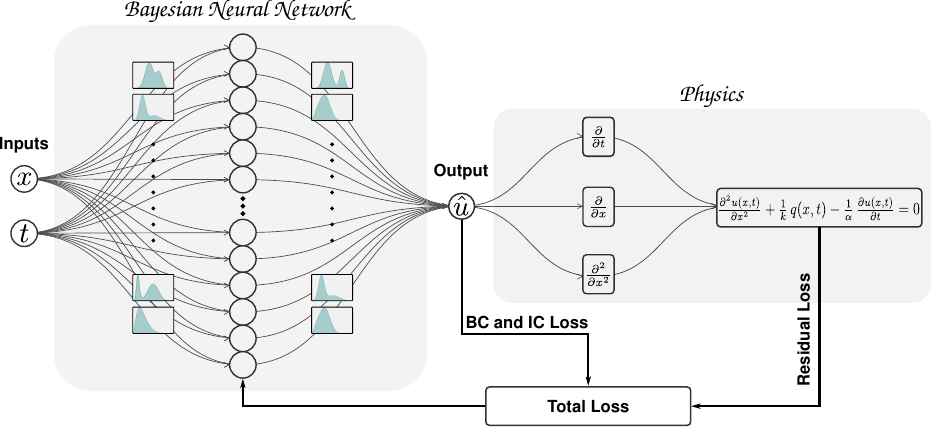}
	\caption{Bayesian-PINN approach with epistemic and homoscedastic uncertainty quantification for the probabilistic spatiotemporal transformer thermal model \cite{BPINN_PHM25}.}
	\label{fig:BPINN_Framework_Homo}
\end{figure}

This section presents the training algorithm of the Bayesian-PINN approach from \cite{BPINN_PHM25}, which integrates homoscedastic and epistemic uncertainty quantification. Figure~\ref{fig:BPINN_Framework_Homo} shows the overall approach and Algorithm~\ref{alg:Bayesian_PINN} outlines the main steps to implement the Bayesian-PINN model using Variational Inference. Further details are provided in \cite{BPINN_PHM25}.

\begin{algorithm}[!htb]
	\caption{B-PINN Training via Variational Inference}
	\label{alg:Bayesian_PINN}
	\begin{algorithmic}[1]
		\State \textbf{Input:} Collocation points $(x_f, t_f)$, initial condition data $(x_0, u_0)$, boundary condition data $(x_{bc}, t_{bc}, u_{bc})$, prior distribution $p(\bm{\theta})$
		\State Initialize variational parameters $\bm{w} = \{\bm{\mu}, \bm{\sigma}\}$ for $\bm{\theta}$
		\While{not converged}
		\State Sample $\bm{\theta}^{(i)} \sim q(\bm{\theta}|\bm{w})$ via reparameterization trick
		\State Evaluate predicted solution: $\hat{u}(x,t;\bm{\theta}^{(i)})$
		\State Compute residual: $r(x_f,t_f;\bm{\theta}^{(i)}) = \mathcal{N}[\hat{u}](x_f, t_f;\bm{\theta}^{(i)}) - f(x_f,t_f)$
		\State Evaluate log-likelihood terms taking the $\log$ of Eq. (\ref{eq:L_BPINN})
		\State Evaluate prior log-probability: $\log p(\bm{\theta}^{(i)})$
		\State Evaluate variational density: $\log q(\bm{\theta}^{(i)}|\bm{w})$
		\State Compute Monte Carlo estimate of ELBO loss via Eq. (\ref{eq:ELBO_BPINN})
		\State Update $\bm{w} = \{\bm{\mu}, \bm{\sigma}\}$ using gradient $\nabla_{\bm{w}} \mathcal{L}^{(i)}$
		\EndWhile
		\State \textbf{Return:} Learned variational posterior $q(\bm{\theta}|\bm{w})$
	\end{algorithmic}
\end{algorithm}

\subsection{Metrics}
\label{ss:Evaluation Metrics}

The accuracy of probabilistic predictions is evaluated using three complementary metrics.

\vspace{2mm} 

\noindent \textbf{Negative Log Likelihood} (NLL) evaluates how well a probabilistic model explains a given set of observations. It is defined as the negative logarithm of the likelihood function, which measures the probability of the observed data under a particular model~\cite{murphy_2012}. For a model with parameters $\bm{\theta}$ and observed data $X\! =\! \{x_1,\! \dots\!,\! x_n\!\}$, NLL is defined as:

\begin{equation}
	\text{NLL}(\theta) = -\sum_{i=1}^n \log p(x_i \mid \bm{\theta})
	\label{eq:nll}
\end{equation}

\noindent where $p(x_i \mid \bm{\theta})$ is the probability of observation $x_i$ given the model parameters. Minimizing the NLL is equivalent to maximizing the likelihood function, as the logarithm is a monotonically increasing function. NLL is particularly useful because it avoids numerical underflow with small probabilities and transforms the product of probabilities into a sum of log probabilities, which is more computationally stable~\cite{bishop_2006}.

\vspace{2mm}

\noindent \textbf{Continuous Ranked Probability Score} (CRPS) measures the discrepancy between the predicted Cumulative Distribution Function (CDF), $F(\cdot)$, and the observed empirical CDF for a given scalar observation $y$~\cite{zamo2018}:

\begin{equation}
	\label{eq:crps}
	CRPS(F,y) = \int (F(x) -\mathds{1}(x\geq y_i))^2 dx,
\end{equation}

\noindent where $\mathds{1}(x\geq y_i)$ is the indicator function, which models the empirical  CDF. 

In order to obtain a single score value from Eq.~(\ref{eq:crps}), a weighted average is computed for each individual observation of the test set~\cite{Gneiting2005}:

\begin{equation}
	\label{eq:crps_avg}
	CRPS = \frac{1}{N} \sum_{i=1}^{N} CRPS(F_i,y_i)
\end{equation}

\noindent where $N$ denotes the total number of predictions.

\vspace{2mm}
\noindent \textbf{Calibration} refers to the statistical consistency between the predictive distributions and the actual observations. It represents a joint property of forecasts and empirical data~\cite{jung2022}. Namely, it is stated that the model is calibrated if~\cite{kuleshov2018}:

    \begin{equation}
        \label{eq:calibration}
       \frac{\sum_{t=1}^{T} \mathds{I}\{y_t \leq F_{t}^{-1}(p)\} }{T} \rightarrow p \text{ for all } p \in [0,1]
    \end{equation}

\noindent In this expression, $T$ refers to the total number of data points, while the indicator function $\mathds{I}\{y_t \leq F_{t}^{-1}(p)\}$ takes a value of 1 when the condition $y_t \leq F_{t}^{-1}(p)$ is true, and 0 otherwise. Given this condition, $y_t$ express the observed outcome at time $t$, and $F_{t}^{-1}(p)$ is the inverse of the CDF for the forecast, evaluated at probability $p$. Therefore, the condition represents the threshold below which a random sample from the distribution would occur with a probability $p$.\\

\vspace{2mm}
\noindent \textbf{Sharpness} means that the confidence intervals should be optimized for minimal width around a singular value. That is, the goal is to reduce the variance, denoted as $var(F_n)$, of the random variable characterized by the cumulative distribution function $F_n$~\cite{kuleshov2018, Tran_2020}:

    \begin{equation}
        \label{eq:sharpness}
         sha = \sqrt{\frac{1}{N} \sum_{n=1}^{N} var(F_n)}
    \end{equation}\\

\newpage

\end{document}